\documentclass[lettersize,journal]{IEEEtran}
\usepackage{amsmath,amsfonts}
\usepackage{algorithmic}
\usepackage{algorithm}
\usepackage{array}
\usepackage[caption=false,font=normalsize,labelfont=sf,textfont=sf]{subfig}
\usepackage{textcomp}
\usepackage{stfloats}
\usepackage{url}
\usepackage{verbatim}
\usepackage{graphicx}
\graphicspath{{figs/}}
\usepackage{cite}
\usepackage{booktabs}
\usepackage{multirow}
\usepackage[table]{xcolor}
\usepackage{pifont}
\usepackage{arydshln}
\newcommand{\cmark}{\ding{51}} % check mark
\newcommand{\xmark}{\ding{55}} % cross mark
\newcommand{\indom}[1]{\cellcolor{gray!15}\textcolor{gray!70}{#1}}
\newcommand{\cropbench}{CropBench}
\newcommand{\cropreward}{CropReward}
\newcommand{\cropdata}{CropReward-300K}
\DeclareRobustCommand{\croprl}{CropAlign}
\newcommand{\croprlsft}{CropAlign\nobreakdash-SFT}

\newcommand{\etal}{\textit{et al.}}
\begin{document}

\title{CropReward: A Human-Preference Benchmark and Reward Model for Aesthetic Image Cropping}

\author{
Xiangfei Sheng,
Xiaofeng Pan,
Leida Li,~\IEEEmembership{Senior Member,~IEEE}, \\
Qian Zhang, 
Biao Wang,
Johan Wagemans,
and Weisi Lin,~\IEEEmembership{Fellow,~IEEE}
% Biao Wang,
% and Johan Wagemans
% and Weisi Lin,~\IEEEmembership{Fellow,~IEEE}
        % <-this % stops a space
\thanks{This work was supported in part by the National Natural Science Foundation of China under Grants 62471349, 625B2142, and 62171340, in part by the Fundamental Research Funds for the Central Universities under Grant QTZX25076. \textit{(Corresponding author: Leida Li)}.

Xiangfei Sheng, Xiaofeng Pan and Leida Li are with the School of Artificial Intelligence, Xidian University, Xi'an 710071, China (e-mails: xiangfeisheng@gmail.com;
panxf@stu.xidian.edu.cn; ldli@xidian.edu.cn).

Qian Zhang and Biao Wang are with Huawei Technologies Co., Ltd., Shenzhen 518129, China (e-mails: qian.zh@huawei.com; wangbiao106@huawei.com).

Johan Wagemans is with the Laboratory of Experimental Psychology, Department of Brain and Cognition, KU Leuven, 3000 Leuven, Belgium (email: johan.wagemans@kuleuven.be).

Weisi Lin is with the College of Computing and Data Science, Nanyang Technological University, Singapore (e-mail: wslin@ntu.edu.sg).

% Weisi Lin is with the College of Computing and Data Science, Nanyang Technological University, Singapore 639798 (e-mail: wslin@ntu.edu.sg).
}% <-this % stops a space
}

% The paper headers
\markboth{}%
{Sheng \MakeLowercase{\textit{et al.}}: CropReward: A Human-Preference Benchmark and Reward Model for Aesthetic Image Cropping}

\maketitle

\begin{abstract}
Aesthetic image cropping (AIC) aims to improve image composition and aesthetic quality by removing extraneous or distracting content. Over the past decade, AIC methods have evolved from specialized vision models to multimodal large language models (MLLMs). Reliable evaluation that reflects human preferences is essential for benchmarking AIC methods and guiding model optimization. However, the current AIC evaluation relies largely on geometry-based metrics such as intersection over union (IoU), as well as image aesthetic assessment(IAA) models. Although inconsistencies with human preferences have long been observed, the robustness and reliability of these evaluation criteria remain underexplored. To bridge this gap, we first construct \cropbench, a large-scale benchmark comprising 106,995 human preference pairs for 1,019 images, with each candidate set containing the original image and candidate crops produced by 14 AIC methods. A systematic study on \cropbench\ reveals that the existing criteria, ranging from IoU to off-the-shelf IAA models, only exhibit limited alignment with human preferences. Building upon \cropbench, we further propose \cropreward, an MLLM-based reward model that evaluates a candidate crop by comparing with the original uncropped image and outputs a score indicating explicit aesthetic improvement or degradation relative to the original. In addition, we employ \cropreward\ to construct \cropdata, the largest AIC dataset to date, and further develop \croprl, a preference-aligned AIC model trained solely with \cropreward\ supervision, without human-annotated crop boxes. Experimental results demonstrate that \cropreward\ aligns with human preferences significantly better than existing AIC evaluation criteria. \croprl\ also surpasses the state-of-the-art AIC methods, demonstrating the advantage of \cropreward\ in guiding AIC model optimization. Models and datasets will be made publicly available.
\end{abstract}

\begin{IEEEkeywords}
Aesthetic image cropping, human preference, benchmark, reward model, image aesthetic assessment.
\end{IEEEkeywords}

\section{Introduction}\label{sec:intro}

\begin{figure}[!t]
\centering
\includegraphics[width=\columnwidth]{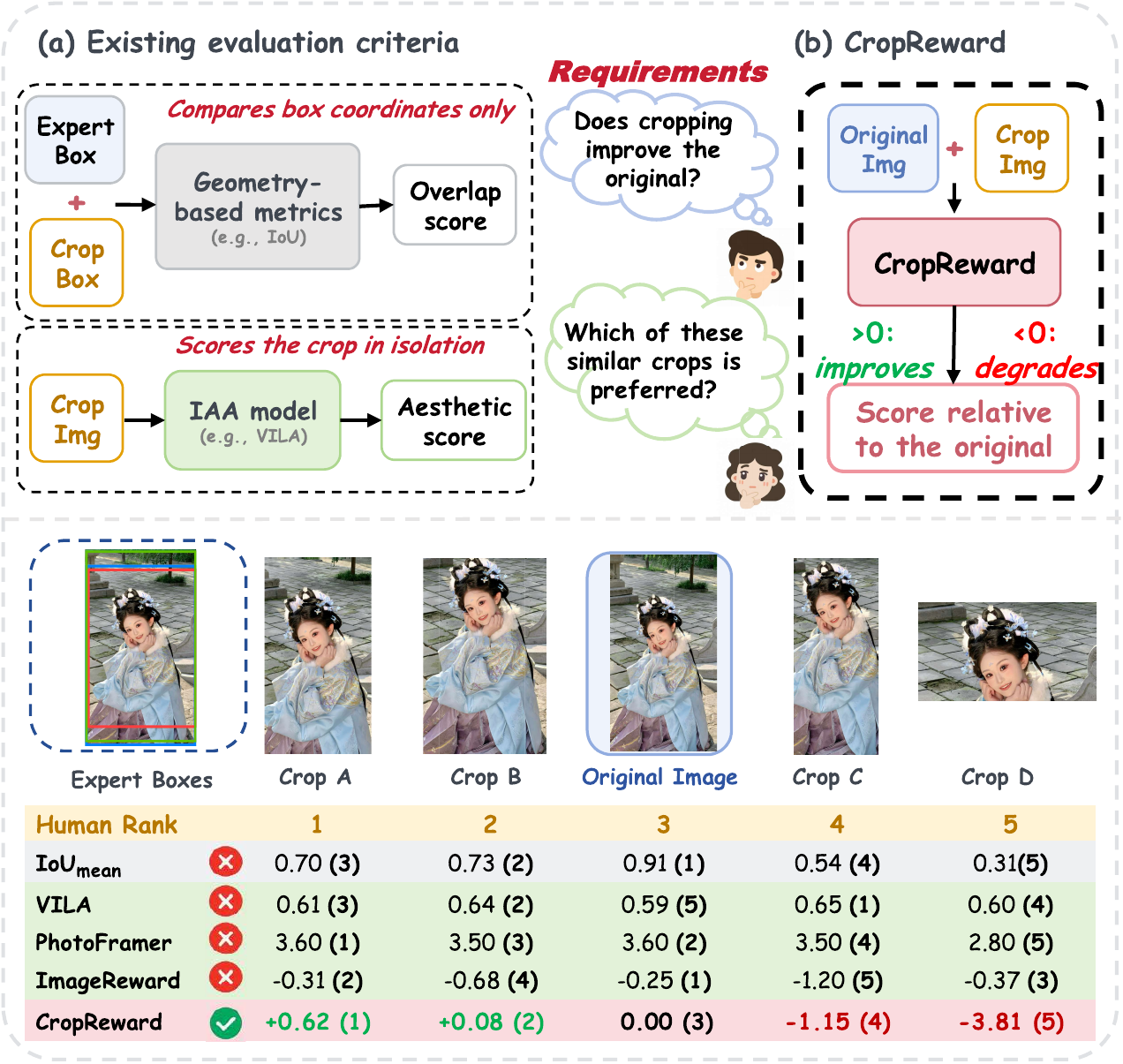}
\caption{Existing evaluation criteria versus the proposed \cropreward.
(a) Geometry-based metrics compare the coordinates of predicted and expert reference boxes, while image aesthetic assessment (IAA) models score each crop in isolation.
(b) \cropreward\ takes the original image and a crop as input and outputs a score relative to the original image, with positive and negative values indicating predicted improvement and degradation, respectively.
\textit{Bottom:} An example from \cropbench. IoU$_{\mathrm{mean}}$ is computed over the three expert reference boxes shown on the left. Numbers in parentheses denote ranks.}
\label{fig:eval-paradigm}
\end{figure}

\IEEEPARstart{A}{esthetic} image cropping (AIC) aims to improve the composition and aesthetic quality of an image by removing extraneous or distracting content~\cite{wang2019cropping,zeng2020gaicd}. AIC has widespread applications in photography, content creation and editing, and thumbnail generation~\cite{yan2013learning,wang2019cropping}. Over the past decade, AIC has made substantial progress through deep learning, with methods evolving from specialized vision models~\cite{zeng2020gaicd,hong2021cacnet} to multimodal large language models (MLLMs)~\cite{lee2025cropper,sheng2025instructcrop,venus2026}. For the same image, different methods may produce multiple plausible crops with subtle differences in content and composition. Reliable evaluation that reflects human preferences over cropping results is therefore essential for benchmarking AIC methods and guiding model optimization.

In the literature, AIC evaluation has relied heavily on geometry-based metrics~\cite{hong2021cacnet,lee2025cropper,dong2026crop}, among which intersection over union (IoU) is the most representative, althouth its inconsistency with human preferences has long been observed~\cite{zeng2020gaicd}. As illustrated in Fig.~\ref{fig:eval-paradigm}(a), metrics like IoU require expert-annotated reference boxes and measure the spatial alignment between a predicted crop and these references. An image may admit multiple pleasing crops with different compositions, so deviating from the reference box does not necessarily indicate a bad crop. Moreover, these metrics only consider box coordinates and may assign similar scores to a crop that preserves the main subject and another one that cuts off part of it. Recent work has also explored using image aesthetic assessment (IAA) models to evaluate cropping results directly~\cite{lee2025cropper,kong2026shotcrop}. However, scoring a crop in isolation does not directly account for what has been removed or how the composition has changed relative to the original image. A high aesthetic score therefore does not necessarily mean that people prefer the crop to the original. In the example at the bottom of Fig.~\ref{fig:eval-paradigm}, VILA~\cite{ke2023vila} ranks crop C first, although annotators prefer the original image to this crop. These limitations motivate us to rethink existing evaluation criteria and raise a central question:

\textit{How should we comprehensively and fairly measure the performance of AIC models?}

We argue that reliable evaluation of AIC methods should meet two requirements. \textbf{(1) Assessing the effectiveness of cropping.} Cropping may improve or degrade the aesthetic quality of the original image. Evaluation should therefore assess whether people prefer the resulting crop to the original image or vice versa. \textbf{(2) Distinguishing subtle differences among candidate crops.} Crops of the same image may differ slightly in their boundaries, yet these differences can affect the content and composition of the resulting crops. Evaluation model should faithfully reflect human preferences among similar crops. However, how well existing evaluation criteria meet the above requirements remains unclear.

To address the above challenges, we first construct \textbf{\cropbench}, a large-scale human preference benchmark for AIC task. It comprises 106,995 human-annotated preference pairs of crops produced by 14 AIC methods on 1,019 images covering both general and portrait photography. The original image is also included when forming the preference pairs, and all candidate pairs are annotated with an explicit option for ties. We further calculate preference scores using a tie-aware Bradley--Terry (BT) model, with the score of the original image anchored at zero within each candidate set. Based on \cropbench, we conduct a systematic study of existing AIC evaluation criteria, ranging from IoU to state-of-the-art IAA models, revealing that all existing criteria achieve only limited consistency with human preferences. \cropbench\ also supports benchmarking of current AIC methods. The results reveal a phenomenon that is often overlooked in AIC: even the top-ranked models sometimes produce crops that people prefer less than the original image.

\begin{figure}[!t]
\centering
\includegraphics[width=\columnwidth]{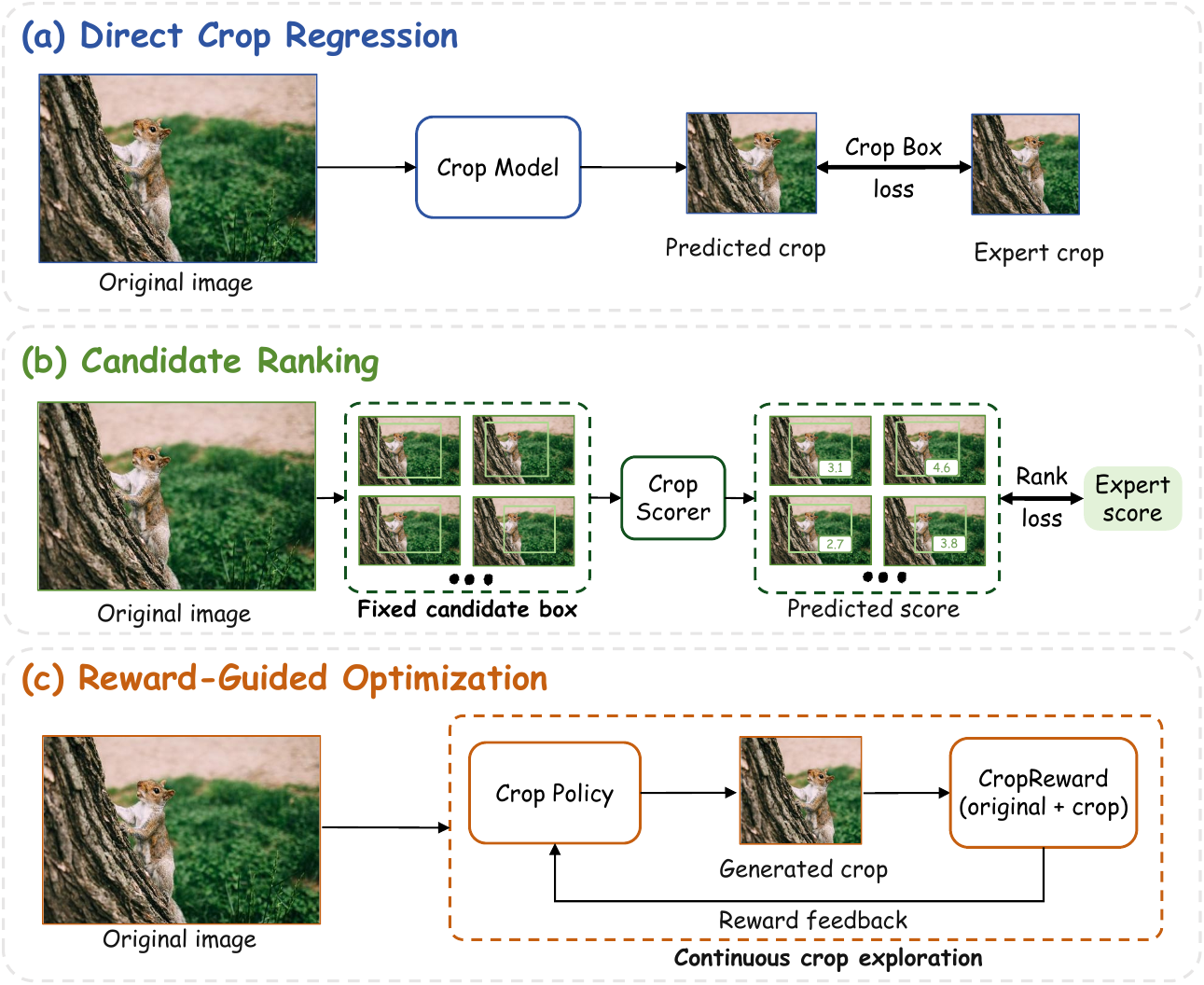}
\caption{Three training paradigms for AIC. (a) Crop regression learns to predict a target crop box. (b) Candidate ranking learns to score candidate crops.
(c) Reward-guided exploration uses \cropreward\ to evaluate each generated crop by comparing with the uncropped original image. The resulting reward is used to optimize the cropping policy toward human preferences without relying on human-annotated crop boxes.}
\label{fig:paradigm-space}
\end{figure}

Building on the above findings, we further propose \textbf{\cropreward}, an MLLM-based reward model trained on the preference annotations of \cropbench, which satisfies the two aforementioned requirements. As illustrated in Fig.~\ref{fig:eval-paradigm}(b), \cropreward\ takes both the original image and a candidate crop as input, allowing it to consider what is retained and what is removed without requiring expert-annotated reference boxes. Its output is expressed relative to the original image, with positive and negative scores indicating predicted improvement and degradation in aesthetic quality, respectively. To distinguish similar crops, we further introduce a pairwise difference learning objective, based on which the model is trained to predict the difference in human preference scores between two crops.
We further explore the potential of \cropreward\ for guiding AIC model optimization. We first use it to select crops generated by existing AIC methods on a large number of unlabeled images, based on which we build \textbf{\cropdata}, a large-scale AIC dataset containing over 300,000 original images, each paired with a CropReward-selected crop box. We then develop \textbf{\croprl}, an MLLM-based cropping model trained through supervised fine-tuning on this dataset and further optimized through reinforcement learning with \cropreward\ as the reward.

\begin{table*}[!t]
  \centering
  \caption{Comparison of \cropbench\ with representative AIC datasets.
  \cmark: included; \xmark: unavailable. Images denote the number of
  source images. ``Original image''
  indicates whether the original image is included as a candidate in
  the comparisons, and ``Explicit pairs'' indicates whether
  candidate-pair labels are included in the public release.}
  \label{tab:crop-dataset-comparison}
  \setlength{\tabcolsep}{3.1pt}
  \renewcommand{\arraystretch}{1.10}
  \resizebox{\textwidth}{!}{%
  \begin{tabular}{llrrllcccl}
    \toprule
    \multirow{2}{*}{\textbf{Dataset}}
      & \multirow{2}{*}{\textbf{Venue}}
      & \multirow{2}{*}{\textbf{Images}}
      & \multirow{2}{*}{\shortstack{\textbf{Candidate}\\\textbf{views}}}
      & \multirow{2}{*}{\textbf{Annotation signal}}
      & \multirow{2}{*}{\shortstack{\textbf{Candidate}\\\textbf{source}}}
      & \multirow{2}{*}{\shortstack{\textbf{Expert}\\\textbf{crops}}}
      & \multirow{2}{*}{\shortstack{\textbf{Original}\\\textbf{image}}}
      & \multirow{2}{*}{\shortstack{\textbf{Explicit}\\\textbf{pairs}}}
      & \multirow{2}{*}{\textbf{Domain}} \\
      & & & & & & & & & \\
    \midrule
    \multicolumn{10}{l}{\textit{Expert-crop datasets}} \\
    ICDB~\cite{yan2013learning} & \textit{CVPR 2013} & 950 & 2,850 & 3 expert crops/image
      & Expert crops & \cmark & \xmark & \xmark & General \\
    FLMS~\cite{fang2014automatic} & \textit{ACM MM 2014} & 500 & 5,000 & 10 expert crops/image
      & Expert crops & \cmark & \xmark & \xmark & General \\
    FCDB~\cite{chen2017quantitative} & \textit{WACV 2017} & 1,743 & --
      & 34,130 pairwise prefs.
      & Expert crops + random subviews & \cmark & \xmark & \cmark & General \\
    \midrule
    \multicolumn{10}{l}{\textit{Candidate-scoring datasets}} \\
    CPC~\cite{wei2018good} & \textit{CVPR 2018} & 10,800 & 259,200
      & Per-annotator candidate labels (0--4)
      & Algorithm-generated + random views & \xmark & \xmark & \xmark & General \\
    GAICD-v1~\cite{zeng2019reliable} & \textit{CVPR 2019} & 1,236 & 106,860 & 5-level MOS
      & Grid anchors & \xmark & \xmark & \xmark & General \\
    GAICD-v2~\cite{zeng2020gaicd} & \textit{TPAMI 2020} & 3,336 & 288,069 & 5-level MOS
      & Grid anchors & \xmark & \xmark & \xmark & General \\
    UGCrop5K~\cite{s2cnet2023} & \textit{AAAI 2024} & 5,000 & 450,000 & 5-level MOS
      & Grid anchors & \xmark & \xmark & \xmark & UGC \\
    \midrule
    \multicolumn{10}{l}{\textit{Synthetic-crop datasets}} \\
    GenCrop~\cite{hong2024gencrop} & \textit{AAAI 2024} & 88K & 290K
      & Outpainting-derived crop pairs
      & Outpainted views & \xmark & \xmark & \xmark & Subject-aware \\
    ProCrop~\cite{zhang2026procrop} & \textit{AAAI 2026} & 242K & 1.936M
      & 8 refined pseudo crops/image
      & Outpainting + refinement & \xmark & \xmark & \xmark & General \\
    \midrule
    \multicolumn{10}{l}{\textit{Fine-grained preference datasets}} \\
    FGAesthetics-Crop~\cite{fgaesq2026} & \textit{CVPR 2026} & 10,554 & --
      & 25,643 pairwise prefs.
      & Crop series & \xmark & \xmark & \cmark
      & General \\
    \midrule
    \rowcolor{gray!13}
    \textbf{\cropbench} & --
      & \textbf{1,019} & \textbf{15,285}
      & \textbf{\begin{tabular}[c]{@{}l@{}}3 expert crops/image\\+ 106,995 pairwise prefs.\\+ 15,285 BT scores\end{tabular}}
      & \textbf{\begin{tabular}[c]{@{}l@{}}10 AIC models + 4 MLLMs\\+ original image\end{tabular}}
      & \cmark & \cmark & \cmark & \textbf{General + Portrait} \\
    \bottomrule
  \end{tabular}%
  }
\end{table*}

The contributions of this work are threefold:
\begin{itemize}

\item We build \textbf{\cropbench}, a large-scale human preference benchmark for AIC task, using a hybrid annotation strategy that determines the number of human annotators per pair based on agreement among five MLLM judges. The benchmark contains 106,995 pairwise labels covering outputs from 14 AIC models and the corresponding original images, supporting systematic evaluation of both existing criteria and AIC methods.
\item We propose \textbf{\cropreward}, an MLLM-based reward model that learns fine-grained cropping preferences to assess whether cropping improves or degrades the original composition, which aligns substantially better with human preferences than existing criteria.
\item We construct \textbf{\cropdata}, a large-scale AIC dataset curated by \cropreward\ without the need of human annotation, and develop \textbf{\croprl}, an advanced cropping model optimized with \cropreward\ that achieves the state-of-the-art performance on multiple benchmarks, including CropBench, and existing cropping datasets, FLMS and FCDB.
\end{itemize}

\section{Related Work}\label{sec:related}

\subsection{Aesthetic Image Cropping}

Early AIC methods searched for crop windows using handcrafted cues such as saliency, composition rules, and boundary simplicity~\cite{fang2014automatic}. Two common paradigms in deep learning-based AIC are \emph{crop regression} and \emph{candidate ranking}, as illustrated in Fig.~\ref{fig:paradigm-space}(a) and Fig.~\ref{fig:paradigm-space}(b). Crop regression methods directly predict a crop box from the input image~\cite{hong2021cacnet,hong2024gencrop,zhang2026procrop}. The composition-aware cropping network (CACNet)~\cite{hong2021cacnet} recognizes composition patterns to guide crop prediction. To alleviate reliance on expert annotations, Hong \etal~\cite{hong2024gencrop} outpainted professional photographs, treating the original image as the target crop of a larger outpainted image. By contrast, candidate ranking methods score a set of candidate crops and select the highest-scoring one~\cite{chen2017vfn,wei2018good,zeng2019reliable,lu2019lvrn,zeng2020gaicd,li2020composing,zhang2022hcic,s2cnet2023}. Wei \etal~\cite{wei2018good} trained a crop scoring network using human judgments of relative crop quality. Zeng \etal~\cite{zeng2019reliable,zeng2020gaicd} proposed grid anchor-based image cropping (GAIC), which uses grid anchors to generate candidate crops and a lightweight model to score them. In crop regression, training against a target box penalizes deviations from that box, even when alternative crops are equally good. Candidate ranking can learn human preferences among crops. However, when the original image is excluded from the training comparisons, these preferences do not establish whether a crop is preferred to the original.

Beyond direct regression and ranking, Li \etal~\cite{li2018a2rl} introduced aesthetics-aware reinforcement learning (A2-RL), an early attempt to optimize cropping through a sequence of window-adjustment actions. Its reward was derived from an image aesthetic classifier rather than learned from human preferences over crop decisions. More recently, MLLMs have been applied to AIC~\cite{lee2025cropper,sheng2025instructcrop,venus2026}. Sheng \etal~\cite{sheng2025instructcrop} applied instruction tuning to MLLMs for AIC. Du \etal~\cite{venus2026} first trained an MLLM to identify aesthetic problems and suggest improvements, then fine-tuned it using chain-of-thought rationales to predict crop boxes and explain the cropping decisions. In this work, we train \cropreward\ on human preferences among candidate crops with the original image included. Then the score of \cropreward\ naturally serves as a better reward for optimizing a cropping model without human-annotated crop boxes, as illustrated in Fig.~\ref{fig:paradigm-space}(c).

\subsection{AIC Datasets and Evaluation Protocols}

Table~\ref{tab:crop-dataset-comparison} summarizes representative AIC datasets categorized by their candidate sources and annotation types. \emph{Expert-crop datasets}~\cite{yan2013learning,fang2014automatic,chen2017quantitative} provide one or several human-annotated crop boxes for each image. These boxes serve as training targets and as references for geometry-based evaluation. \emph{Candidate-scoring datasets}~\cite{wei2018good,zeng2019reliable,zeng2020gaicd,s2cnet2023} collect human judgments of crop quality over candidates generated by existing algorithms, grid anchors, or random sampling. These annotations support training and evaluation of candidate ranking models. \emph{Synthetic-crop datasets}~\cite{hong2024gencrop,zhang2026procrop} construct training pairs automatically through outpainting and pseudo-crop refinement. FGAesthetics~\cite{fgaesq2026} collects pairwise preferences among visually similar images for fine-grained aesthetic assessment. Its cropping subset contains 1,702 series derived from CPC and GAICD datasets, comprising 10,554 crops and 25,644 preference pairs.

Evaluation protocols depend on the types of annotations. 
IoU and boundary displacement error (Disp) measure the geometric alignment between predicted and reference crop boxes, whereas ranking metrics such as $\mathrm{Acc}_{K/N}$ and SRCC compare the predicted scores of candidate crops with their mean opinion scores (MOS)~\cite{zeng2020gaicd}.
However, these protocols do not accurately assess whether people prefer the predicted crops relative to the original image or vice versa. \cropbench\ comprises the crops of diverse AIC models and the original image in each candidate set, with human preference annotations for all candidate pairs and an explicit option for ties. Expert-annotated crop boxes are also provided for computing geometric metrics. These annotations support various AIC evaluation tasks, including benchmarking AIC methods, evaluating how well existing criteria reflect human preferences, and training a reward model.

\subsection{Image Aesthetic and Composition Assessment}

IAA predicts the aesthetic quality of an image, typically as a score or a distribution of human ratings~\cite{talebi2018nima}. To preserve composition and fine details during feature extraction, some methods adopt multi-scale image representations~\cite{ke2021musiq,behrad2025charm}. Beyond visual features alone, vision-language-based methods learn aesthetic representations from images paired with aesthetic comments~\cite{sheng2023aesclip,ke2023vila}, and MLLM-based models go a step further by producing natural-language explanations of aesthetic judgments~\cite{liao2025humanaesexpert,cao2026artimuse}. More recently, learning from relative rankings has been explored to distinguish aesthetic differences among similar images~\cite{fgaesq2026}.

Image composition assessment (ICA) focuses on the spatial arrangement of visual elements. Composition scoring methods model salient regions and their arrangement under different composition patterns~\cite{zhang2021sampnet}. Recent MLLM-based work further explores explanations of composition quality and guidance for improving composition~\cite{you2025photoframer,sha2026portraitcraft}. 

IAA and ICA models have also been used to evaluate cropping results~\cite{lee2025cropper,kong2026shotcrop}. However, the discriminative ability of the existing IAA/ICA models are typically limited. In addition, an absolute aesthetic or composition score does not directly indicate whether a crop is preferred to the original image. In this work, we propose \cropreward\ to learn this kind of comparison from human cropping preferences, using both the original image and the crop as input.

\subsection{Reward Models for Visual Preference Alignment}
Reward models learn to predict which outputs people prefer and can be used to evaluate generative models and optimize them through reinforcement learning. They have been studied in text-to-image generation~\cite{xu2023imagereward,kirstain2023pick,wu2023hpsv2,ma2025hpsv3}, with subsequent applications to image editing~\cite{editreward2026,luo2026editscore}, photo retouching~\cite{beautygrpo2026}, and image restoration~\cite{liu2026llbench}. In this work, \cropreward\ is trained on human preferences among candidate crops and between crops and their original uncropped images. It evaluates each crop together with the original image and provides a positive/negative score relative to the original, which can serve as both an evaluation criterion and a reward for AIC model optimization.

\begin{figure*}[!t]
\centering
\includegraphics[width=\textwidth]{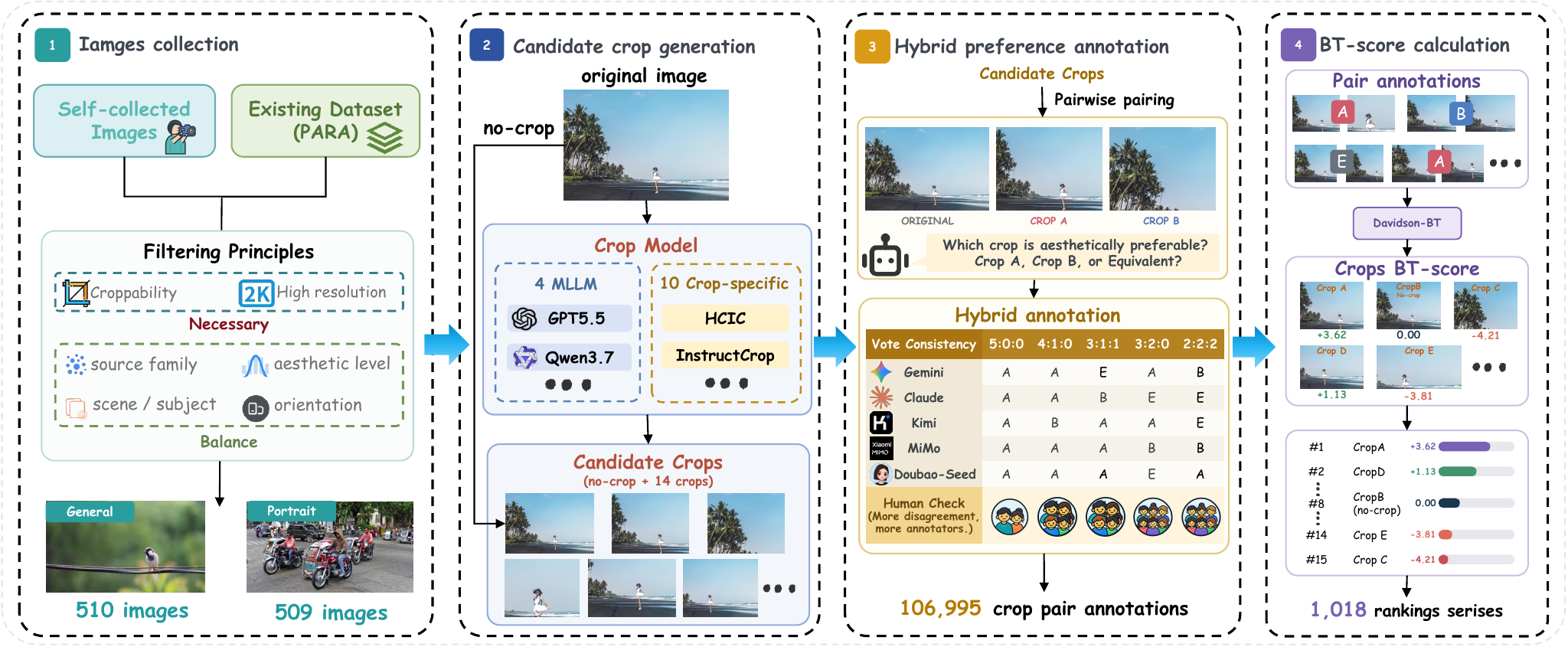}
\caption{Four-stage construction pipeline of \cropbench: image collection, candidate crop generation, preference annotation, and BT-score calculation. Each image has 14 model-generated crops and the original uncropped image as a candidate crop set. Agreement among five MLLM judges determines the number of human annotators assigned to each pair. BT scores are calculated from the final human preference labels, with the score of the original uncropped image anchored at zero.}
\label{fig:cropbench-pipeline}

\par\vspace{6pt}

\centering
\includegraphics[width=\textwidth]{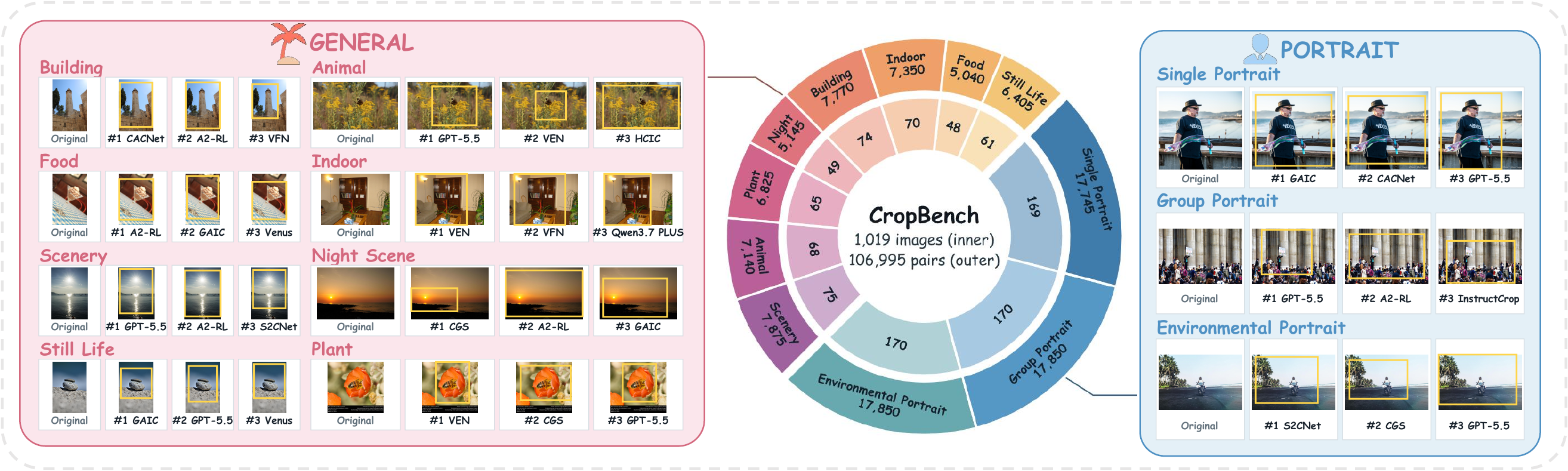}
\caption{Overview and statistics of \cropbench. The sunburst chart reports the number of source images (inner ring) and pairwise comparisons (outer ring) for each category, where the portrait domain (left, blue) covers single, group, and environmental portraits, and the general domain (right, pink) covers eight scenarios. For each example, the original image and three representative candidates are shown, ranked by the derived BT scores.}
\label{fig:cropbench-overview}
\end{figure*}

% CropBench
% - Source images: 1,019
%   - General: 510
%   - Portrait: 509
% - Model-generated crops: 14,266
% - No-crop candidates: 1,019
% - Candidate views: 15,285
% - Pairwise comparisons: 106,995
% - Train: 819 sources / 12,285 candidates / 85,995 pairs
% - Test: 200 sources / 3,000 candidates / 21,000 pairs

\section{\cropbench: A Human-Preference Benchmark for AIC}\label{sec:cropbench}

\cropbench\ provides human preference annotations for benchmarking AIC methods, which is the foundation for assessing existing evaluation criteria and training a reward model. It consists of candidate crop sets and each candidate set contains both the outputs of AIC models and the original uncropped image. We annotate all candidate pairs with an explicit option for ties, allowing comparisons both among crops and with the original image. Fig.~\ref{fig:cropbench-pipeline} illustrates the four stages of dataset construction, and Fig.~\ref{fig:cropbench-overview} shows examples of candidate crop sets from different categories, along with the number of sets in each category.

\subsection{Image Collection}\label{sec:image-collection}
CropBench covers images from both general scenes and portrait photography. For general scenes, eight common categories are considered, including buildings, animals, food, indoor scenes, scenery, night scenes, still life, and plants. For portrait photography, single, group, and environmental portraits are included. Images are drawn from the PARA dataset~\cite{yang2022para}, supplemented by 160 self-collected portrait photographs. We focus on images whose composition has potential for improvement through cropping. An MLLM is first filters out that are already well-composed, would benefit only marginally, or would be harmed by cropping. Three aesthetic experts then review the remaining images. This screening removes 50.4\% of the initial general pool and 55.0\% of the portrait pool.

We sample the remaining images across aesthetic levels, image orientations, and scene/subject categories. Three aesthetic levels are defined using the ratings provided by PARA for general images and HumanAesExpert~\cite{liao2025humanaesexpert} scores for portraits. The final benchmark contains 510 general images and 509 portrait images, with median shorter-side lengths of 1,475 and 1,627 pixels, respectively.

\subsection{Candidate Crop Generation}\label{sec:candidate-generation}
The candidate crop pool is built from the outputs of real cropping models that span the paradigm spectrum of Sec.~\ref{sec:related}. Specifically, we deploy 10 specialized AIC models, including the regression-based CACNet~\cite{hong2021cacnet}, the ranking-based VFN~\cite{chen2017vfn}, VEN~\cite{wei2018good}, GAIC~\cite{zeng2020gaicd}, HCIC~\cite{zhang2022hcic}, CGS~\cite{li2020composing}, and S2CNet~\cite{s2cnet2023}, the RL-based A2-RL~\cite{li2018a2rl}, and the MLLM-based InstructCrop~\cite{sheng2025instructcrop} and Venus~\cite{venus2026}, together with 4 general-purpose MLLMs, namely GPT-5.5~\cite{openai2026gpt55}, Qwen3.5-4B~\cite{qwen2026qwen35}, Qwen3.6-27B~\cite{qwen2026qwen36}, and Qwen3.7-Plus~\cite{qwen2026qwen37plus}. 
Each model contributes one crop per image. All specialized AIC models are applied with their officially released weights. For candidate ranking methods, we select the top-ranked crop; for other methods, we use the crop they produce directly. The four general-purpose MLLMs use the same prompt, which requests a single crop box in normalized coordinates. Including the original uncropped image produces 15 candidates per image and 15,285 candidates in total. Comparing all $\binom{15}{2}=105$ pairs per image yields 106,995 pairs for annotation.

\subsection{Hybrid Preference Annotation}\label{sec:hybrid-annotation}
Collecting human preferences for 106,995 candidate pairs requires substantial annotation effort. To address this challenge, we design a human-machine collaborative annotation (HMCA) paradigm. Specifically, we use MLLM votes to determine how many human annotators assess each pair, while humans provide all final labels. For each comparison, the judge is shown the original image followed by two candidates and asked to choose candidate A, candidate B, or equivalent (E). Five MLLMs first judge every pair independently, including Gemini 3.5 Flash~\cite{google2026gemini35}, Claude Opus 4.8~\cite{anthropic2026claude48}, Doubao-Seed-2.0-Pro~\cite{seed2026seed20}, Kimi K2.5~\cite{team2026kimi}, and MiMo-V2.5~\cite{xiaomi2026mimov25}. These judges are different from the candidate-generating models in Sec.~\ref{sec:candidate-generation}. Agreement among the five MLLM votes determines the number of human annotators assigned to each pair. Pairs with unanimous votes (5:0:0) are assigned to two annotators, pairs with vote counts of 4:1:0 or 3:1:1 to four annotators, and pairs with vote counts of 3:2:0 or 2:2:1 to six annotators. Human annotators do not see the MLLM votes. If the assigned human annotators disagree, an additional annotator reviews the pair in a second round. Final labels are determined by majority voting. All 106,995 pairs are annotated by a pool of twenty graduate students specializing in image aesthetics and composition assessment.

We additionally collect three expert-annotated crop boxes per image for geometric evaluation. Three annotators independently select one crop box for each image, aiming to improve overall composition while preserving important visual content and removing distracting regions. The annotation is performed directly on the original images, without referring to model-generated candidates. These reference boxes are separate from the 15 candidates used for preference annotation.

\subsection{BT-score Calculation}\label{sec:bt-estimation}
We calculate a preference score for each candidate crop from the pairwise annotations. Let $\{c_0, c_1, \ldots, c_{14}\}$ denote the candidate crop set of an image, where $c_0$ denotes the original image and $u_i$ is the preference score of $c_i$, $i=0,1,2,\ldots,14$. To account for ties, we use the Davidson extension of the Bradley--Terry model~\cite{bradley1952rank,davidson1970ties}. For a pair of crops $(c_i, c_j)$, the model defines
\begin{equation}\label{eq:davidson}
P(c_i \succ c_j) = \frac{e^{u_i}}{Z_{ij}}, \qquad
P(c_i \equiv c_j) = \frac{2\nu\, e^{(u_i + u_j)/2}}{Z_{ij}},
\end{equation}
where $Z_{ij} = e^{u_i} + e^{u_j} + 2\nu\, e^{(u_i + u_j)/2}$ and $\nu \geq 0$ controls the probability of a tie, and $\equiv$ denotes an equivalent preference between two candidate crops. Given $\nu$, we estimate the scores separately for each image by maximizing the regularized log-likelihood of its 105 annotated outcomes,
\begin{equation}\label{eq:mle}
\max_{\{u_i\}} \;\; \sum_{(i,j)} \log P\big(y_{ij} \,|\, u_i, u_j, \nu\big) \;-\; \frac{1}{2\sigma^2} \sum_{i \neq 0} u_i^2, \quad \text{s.t.}\;\; u_0 = 0,
\end{equation}
where $y_{ij}$ is the annotated label. The Gaussian prior with $\sigma=3$ limits extreme scores for candidate crops that consistently win or lose. We anchor the score of the original image at zero, so that positive and negative scores indicate improvement and degradation relative to the original image, respectively. \cropbench\ is split by image into a training set of 819 images (85,995 pairs) and a test set of 200 images (21,000 pairs). All candidate crops and comparisons from an image belong to the same split.

\begin{figure*}[!t]
\centering
\includegraphics[width=0.85\textwidth]{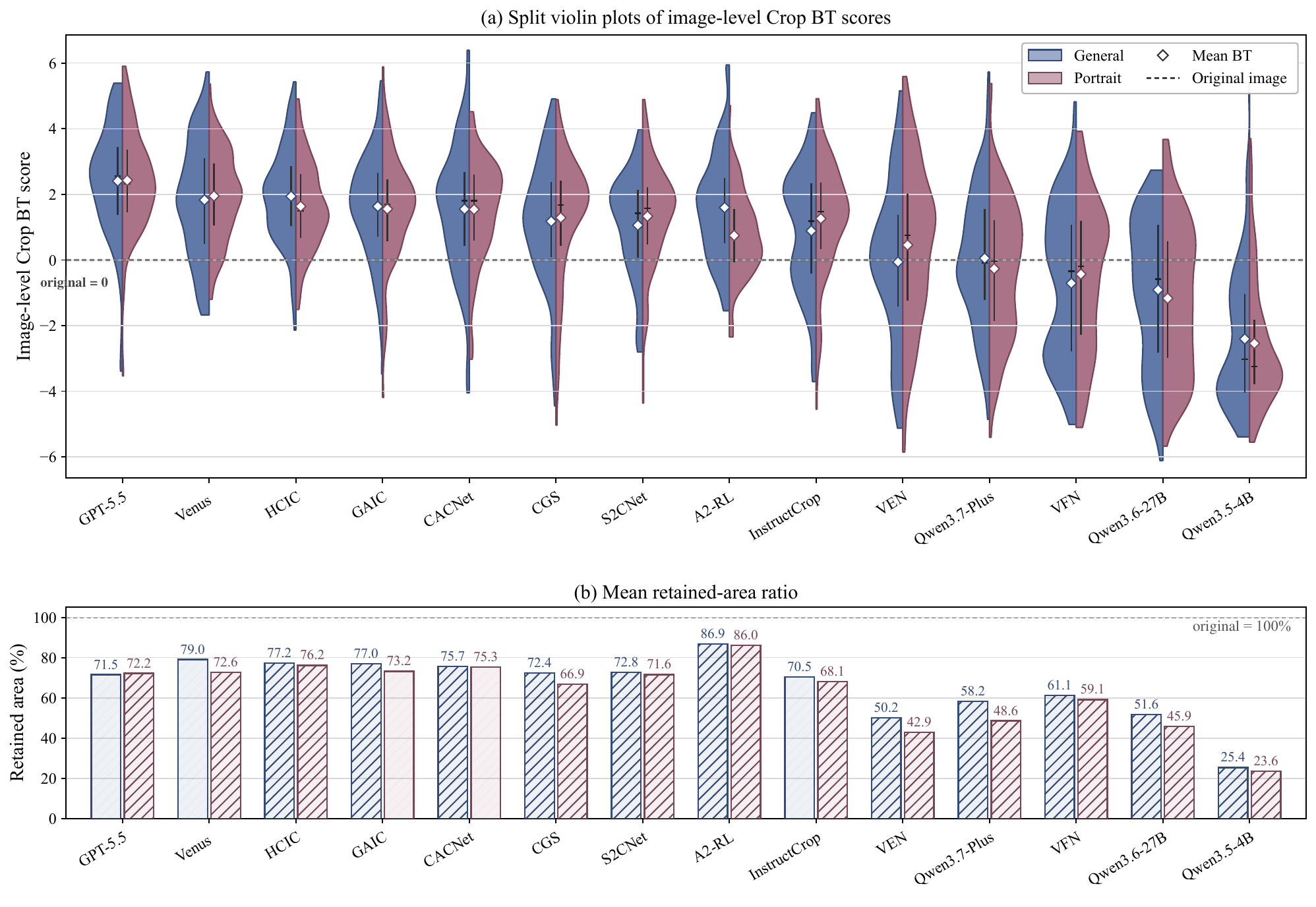}
% AUTHOR CHECK: Verify that the vertical bars are 95% bootstrap confidence
% intervals for the mean BT scores, rather than another measure of spread.
\caption{Performance and cropping extent of the 14 AIC models on \cropbench.
(a) Distributions of per-image BT scores for general (blue) and portrait
(purple) images, with methods ordered by their overall mean scores. Diamonds
denote mean BT scores, and the dashed line marks the original image score of zero.
(b) Mean retained-area ratio, defined as the crop area divided by the
original image area.}
\label{fig:bt-analysis}
\end{figure*}

\subsection{Benchmark Analysis}\label{sec:benchmark-analysis}

We compare the 14 AIC models using their BT scores. Fig.~\ref{fig:bt-analysis}(a) shows the score distributions for general and portrait images. To examine the relationship between cropping extent and human preferences, Fig.~\ref{fig:bt-analysis}(b) shows the mean retained-area ratio for each AIC model. We have three observations.

\textbf{Observation 1: A general-purpose MLLM outperforms specialized AIC models.} GPT-5.5 achieves the highest mean BT score for both image categories under zero-shot evaluation, exceeding all ten evaluated specialized AIC models. Venus achieves the second-highest mean score. Significant performance variations have also been observed across MLLMs. These results motivate us to study whether human preference supervision can further improve AIC models.

\textbf{Observation 2: Cropping does not always improve the composition of the original image.} Several methods have negative mean BT scores. Even the highest-ranked methods may produce some crops with negative scores. These results highlight the need to assess a crop relative to its original, even when the method performs well on average.

\textbf{Observation 3: Methods differ substantially in cropping extent.} A2-RL retains an average of 86.9\% of the original image area for general images and 86.0\% for portraits, whereas Qwen3.5-4B retains only 25.4\% and 23.6\%, respectively. GPT-5.5 achieves higher mean BT scores than the above two methods while retaining approximately 72\% of the image area for each image category. Retained area alone is thus not informative about crop quality, which also depends on what is kept and how it is composed.

\section{\cropreward: A Human-Aligned Reward Model for AIC}\label{sec:cropreward}

\begin{figure*}[!t]
\centering
\includegraphics[width=0.95\textwidth]{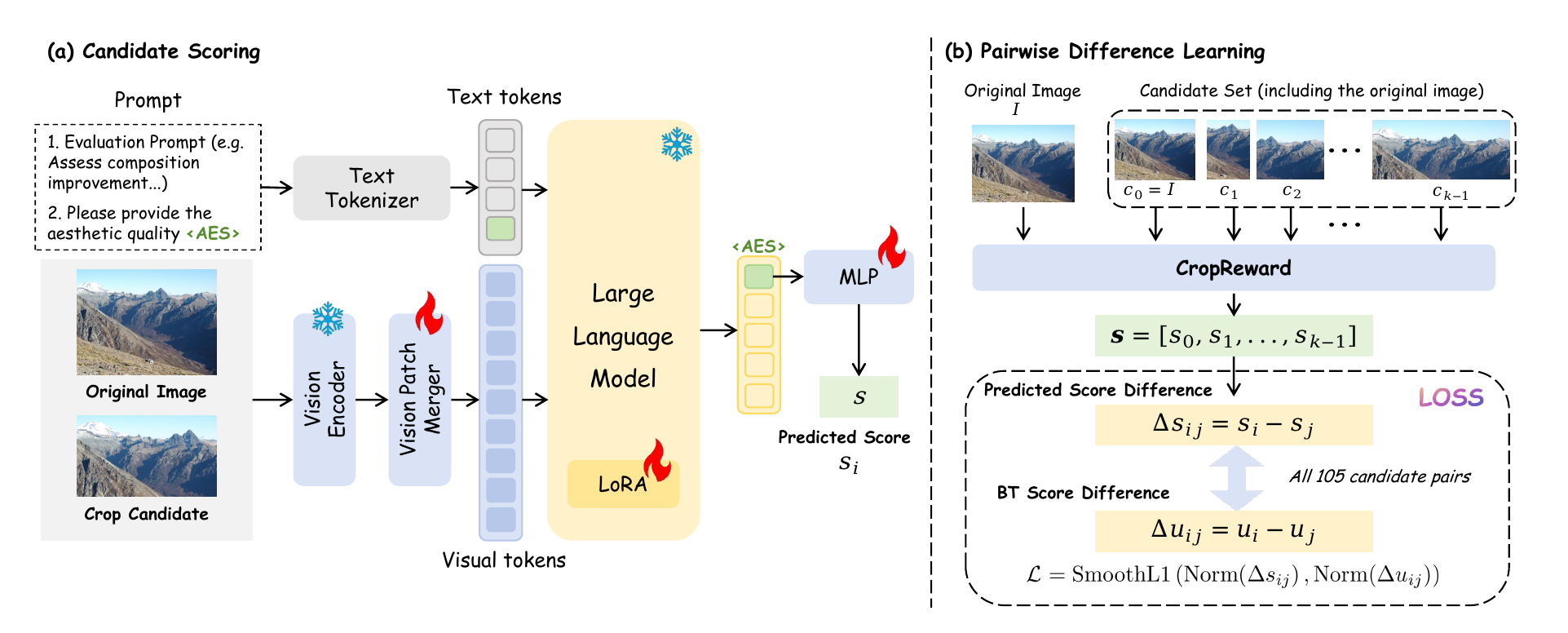}
\caption{Framework of \cropreward. (a) Candidate scoring: an MLLM takes
the original image and one candidate crop as input. A regression head
predicts a score from the final-layer representation of the
\texttt{<AES>} token. (b) Pairwise difference learning: the model is trained
to match predicted score differences to BT preference score differences
over all candidate pairs, including comparisons with the original image.}
\label{fig:cropreward}
\end{figure*}

\subsection{Formulation}\label{sec:formulation}

\cropreward\ learns a scoring function $s_\theta(I,c)$ that takes an original image $I$ and a candidate crop $c$ as input and returns a scalar score. For each training image, let $\mathcal{C}_I=\{c_0,c_1,\ldots,c_{14}\}$ denote the corresponding candidate crop set, where $c_0=I$ is the original image. Each candidate $c_i$ has a BT preference score $u_i$ estimated from human annotations, with $u_0=0$ as defined in Sec.~\ref{sec:bt-estimation}. We train the model to match differences in these preference scores:
\begin{equation}\label{eq:goal}
s_\theta(I,c_i)-s_\theta(I,c_j)
\;\approx\; u_i-u_j,
\qquad 0\leq i<j\leq14.
\end{equation}

\indent This objective includes comparisons among candidate crops and between each crop and the original image. We implement it using the normalized pairwise difference (NPD) loss in Sec.~\ref{sec:diff-learning}. Expert-annotated reference crop boxes are not required for training or inference.

\subsection{Model Architecture}\label{sec:architecture}
We build \cropreward\ on Qwen3.5-4B~\cite{qwen2026qwen35}. As illustrated in Fig.~\ref{fig:cropreward}(a), each candidate crop is scored independently by comparing with the original image with an evaluation prompt. The prompt asks the model to assess composition improvement and content preservation. We append a learnable \texttt{<AES>} token to the prompt and use its final-layer hidden state, denoted by $\mathbf{h}_i$, to predict the candidate score through a two-layer MLP head $\mathcal{R}_\omega$:
\begin{equation}\label{eq:score}
s_i = s_\theta(I,c_i) = \mathcal{R}_\omega(\mathbf{h}_i).
\end{equation}

We train the visual patch merger, low-rank adaptation (LoRA) adapters with rank 16 in the language model, the \texttt{<AES>} embedding, and the MLP head. The remaining pretrained parameters are frozen, giving 62M trainable parameters in total.

\subsection{Pairwise Difference Learning}\label{sec:diff-learning}
To implement the learning objective in Eq.~\eqref{eq:goal}, we use the NPD loss to compare predicted and BT score differences over all candidate pairs, as illustrated in Fig.~\ref{fig:cropreward}(b):
\begin{equation}\label{eq:loss}
\mathcal{L} = \frac{S}{|\mathcal{P}|}\sum_{(i,j)\in\mathcal{P}}
\ell_{\delta}\!\left(\frac{s_i - s_j}{S},\; \frac{u_i - u_j}{S}\right),
\end{equation}
where $\mathcal{P}$ contains all $\binom{15}{2}=105$ unordered pairs, $\ell_{\delta}$ is the smooth-$\ell_1$ loss with transition parameter $\delta$, and $S>0$ is a scale estimated from the training set and fixed throughout training. The transition point $\delta$ is defined in normalized units.

Unlike a ranking loss that uses only preference order, NPD also supervises the difference in BT preference scores. All candidate pairs contribute to the loss, including those annotated as equivalent. We train \cropreward\ jointly on general and portrait images from the \cropbench\ training split. Optimization settings are provided in Sec.~\ref{sec:exp-setup}.

\subsection{Inference}\label{sec:inference}
The NPD loss supervises differences between candidate scores. At inference, we score the original image as a candidate, obtaining $s_\theta(I,I)$. The final score of a crop is computed relative to the original image:
\begin{equation}\label{eq:relative-score}
r_\theta(I,c) = s_\theta(I,c) - s_\theta(I,I),
\end{equation}
where $c$ is a candidate crop of the original image $I$.

A positive score predicts a preference for the crop, while a negative score predicts a preference for the original image. Subtracting the original image score preserves the ranking of candidates from the same image.

\section{Optimizing AIC Models with \cropreward}\label{sec:cropdata}

In this section, we use \cropreward\ to select training crops from a large number of unlabeled images and to optimize a cropping model through reinforcement learning.

% Use independent column floats so text can follow Fig. 7 directly.
\edef\cropdatasavedtopnumber{\number\value{topnumber}}
\setcounter{topnumber}{1}
\suppressfloats[t]
\begin{figure}[t]
\centering
\setlength{\abovecaptionskip}{0pt}
\includegraphics[width=\columnwidth]{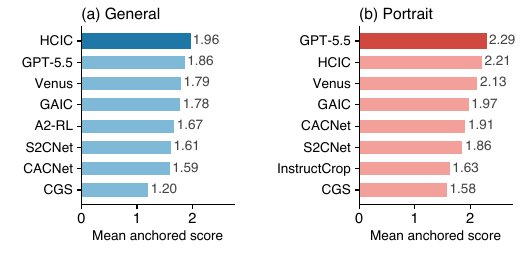}
\caption{Mean \cropreward\ scores of the eight AIC models used to
generate candidates in each domain, computed on the corresponding
general or portrait images used to construct \cropdata. Scores are computed
relative to the original image.}
\label{fig:cropdata-stats}
\vspace{-14pt}
\end{figure}

\subsection{Construction of \cropdata\ Dataset}\label{sec:cropdata-construction}
To obtain training data without additional human annotation, we use \cropreward\ to select crops generated by existing AIC models on unlabeled images. The selected crops serve as labels for \cropdata. We start with 517,889 Unsplash images, comprising 249,797 general images and 268,092 portrait images. For each of the general and portrait domains, we selected the eight top-ranked AIC models on \cropbench\ to generate eight candidate crops per image. We evaluate these eight crops using the relative score $r_\theta(I,c)$ defined in Eq.~\eqref{eq:relative-score} and select the highest-scoring crop as the pseudo label. We then rank images by the scores of their selected crops and retain the top 60\% within each domain.
The resulting dataset contains 310,735 images, including 149,879 general images and 160,856 portraits. Each image is paired with the selected crop, its \cropreward\ score, and the method that produced it. Fig.~\ref{fig:cropdata-stats} shows the mean \cropreward\ scores of the eight models in each domain.

\subsection{Reward-Guided Cropping}\label{sec:reward-guided-cropping}
\begin{figure}[t]
\centering
\setlength{\abovecaptionskip}{0pt}
\includegraphics[width=\columnwidth]{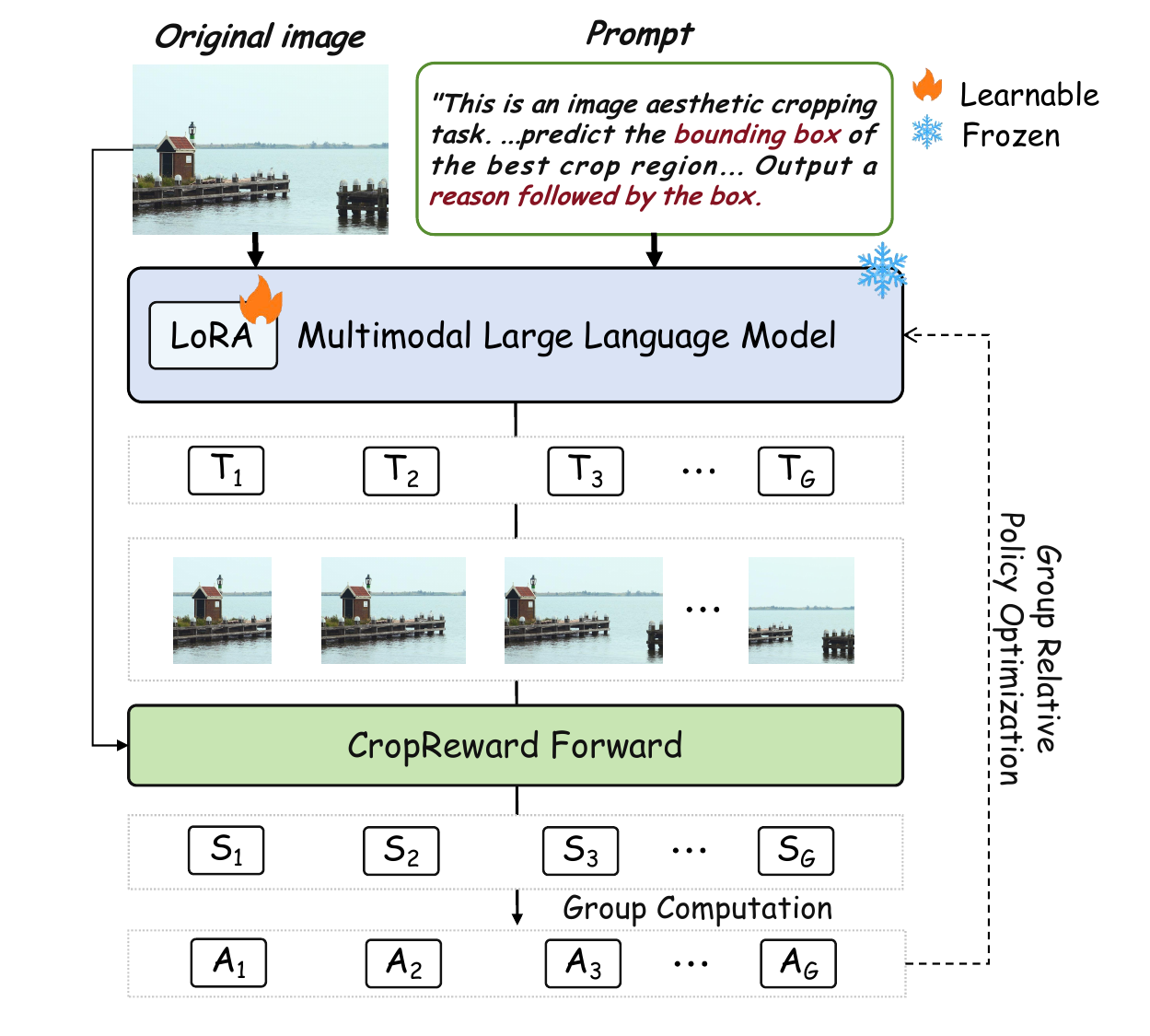}
\caption{Reward-guided cropping with \cropreward. The LoRA-adapted
policy generates a group of $G$ responses containing a textual rationale
followed by a crop box. The frozen \cropreward\ scores each valid crop
relative to the original image. These scores are used to construct
rewards, which are normalized within each group to compute advantages
for the GRPO update.}
\label{fig:grpo}
\vspace{-14pt}
\end{figure}
\setcounter{topnumber}{\cropdatasavedtopnumber}

Based on \cropdata, we first apply supervised fine-tuning (SFT) to Qwen3.5-4B to predict a single crop box in normalized coordinates, producing \croprlsft. We then train \textbf{\croprl} by optimizing \croprlsft\ with group relative policy optimization (GRPO)~\cite{shao2024deepseekmath}, using \cropreward\ to evaluate the crops it generates, as illustrated in Fig.~\ref{fig:grpo}. During this stage, the prompt asks for a textual rationale followed by a crop box, and \cropreward\ remains frozen.

Given an original image $I$ and a task prompt $q$, the policy $\pi_\phi$ samples $G$ responses $\{o^{(1)},\ldots,o^{(G)}\}$. A response is valid if it follows the required output format and contains coordinates defining a valid crop of $I$. For each valid response, we extract the box $b^{(i)}$ and apply it to $I$ to obtain the crop $c^{(i)}$. We use the relative score from \cropreward\ to construct the reward:
\begin{equation}\label{eq:grpo-reward}
R^{(i)} =
\begin{cases}
    \sigma\!\left(r_\theta(I,c^{(i)})\right) + \lambda_f, & \text{if } o^{(i)} \text{ is valid},\\[2pt]
0, & \text{otherwise},
\end{cases}
\end{equation}
where $\sigma(\cdot)$ is the sigmoid function, which maps the relative crop score to $(0,1)$, and $\lambda_f$ is a format bonus. Rewards are normalized within each group to compute relative advantages. GRPO then updates the policy using a clipped policy-gradient objective with a KL penalty relative to a frozen reference policy. The RL stage optimizes the policy with these rewards without target crop boxes, as illustrated in Fig.~\ref{fig:paradigm-space}(c). Neither stage requires additional human annotation.

\section{Experiments}\label{sec:experiments}

In this section, we conduct extensive experiments and comparisons on \cropbench\ and popular AIC datasets to validate the performance of the proposed model from the following four aspects.
\begin{itemize}
\item \textbf{Alignment with Human Preferences:} We evaluate how well existing AIC evaluation criteria agree with human cropping preferences and whether \cropreward\ achieves better alignment (see Sec.~\ref{sec:exp-agreement}).
\item \textbf{Generalization:} We examine whether \cropreward\ generalizes to datasets beyond \cropbench\ (see Sec.~\ref{sec:exp-generalization}).
\item \textbf{Ablation Studies:} We examine the effects of the training objective and the inclusion of the original image as input on the performance of \cropreward\ (see Sec.~\ref{sec:exp-ablation}).
\item \textbf{Effectiveness of Reward:} We assess the effectiveness of \cropreward\ in guiding AIC model optimization by comparing \croprl\ with state-of-the-art AIC models (see Sec.~\ref{sec:exp-downstream}).
\end{itemize}

\subsection{Experimental Setup}\label{sec:exp-setup}

\textbf{Datasets.} We use five datasets for evaluation.
\begin{itemize}
\item \textbf{\cropbench} is used to evaluate agreement with human cropping preferences. It contains 819 training images and 200 test images (100 per domain), with 85,995 and 21,000 annotated pairs, respectively. We report agreement results on the test split.
\item \textbf{GAICD-v2}~\cite{zeng2020gaicd} and \textbf{FGAesthetics-Crop}~\cite{fgaesq2026} are used to evaluate model generalization. We use the 500-image test split of GAICD-v2. FGAesthetics-Crop is the cropping subset of FGAesthetics, and we use the official test split, which contains 168 image series, 990 candidate crops, and 2,088 preference pairs.
\item \textbf{FLMS}~\cite{fang2014automatic} and \textbf{FCDB}~\cite{chen2017quantitative}, together with the test set of \cropbench, are used to evaluate cropping models. We use all 500 FLMS images, with up to ten expert crops per image, and the 345 valid FCDB test images, each with a single expert crop.
\end{itemize}

\begin{table*}[!t]
  \centering
  \caption{Agreement with human cropping preferences on the
  \cropbench\ test split. For each candidate, IoU is computed against
  three expert crop boxes. The subscripts $\min$, $\mathrm{mean}$, and
  $\max$ indicate the minimum, mean, and maximum of these three values,
  respectively. Bold indicates the best result
  within each group in each column.}
  \label{tab:crop-metric-benchmark}
 
  \setlength{\tabcolsep}{2.6pt}
  \renewcommand{\arraystretch}{1.10}
  \resizebox{\textwidth}{!}{%
  \begin{tabular}{llc@{\hspace{5pt}}ccccc@{\hspace{5pt}}ccccc}
    \toprule
    \multirow{2}{*}{\textbf{Method}}
      & \multirow{2}{*}{\textbf{Venue}}
      & \multirow{2}{*}{\textbf{\#Params (B)}}
      & \multicolumn{5}{c}{\textbf{General}}
      & \multicolumn{5}{c}{\textbf{Portrait}} \\
    \cmidrule(lr){4-8}\cmidrule(lr){9-13}
      & & & $\mathrm{PLCC}$ & $\mathrm{SRCC}$
      & $\mathrm{ACC}_{\mathrm{pair}}$
      & $\mathrm{ACC}_{\mathrm{top1}}$
      & $\mathrm{ACC}_{\mathrm{top3}}$
      & $\mathrm{PLCC}$ & $\mathrm{SRCC}$
      & $\mathrm{ACC}_{\mathrm{pair}}$
      & $\mathrm{ACC}_{\mathrm{top1}}$
      & $\mathrm{ACC}_{\mathrm{top3}}$ \\
    \midrule
    \multicolumn{13}{l}{\textbf{Geometry-based metrics} \textit{(three expert crop references)}} \\
    IoU$_{\min}$
      & -- & \textit{n/a}
      & 0.577
      & 0.509
      & 0.672
      & 0.190
      & 0.390
      & 0.684
      & 0.609
      & 0.703
      & 0.200
      & 0.430 \\
    IoU$_{\mathrm{mean}}$
      & -- & \textit{n/a}
      & \textbf{0.684}
      & \textbf{0.623}
      & \textbf{0.714}
      & \textbf{0.210}
      & \textbf{0.490}
      & \textbf{0.755}
      & \textbf{0.657}
      & \textbf{0.721}
      & \textbf{0.230}
      & \textbf{0.480} \\
    IoU$_{\max}$
      & -- & \textit{n/a}
      & 0.674
      & 0.598
      & 0.705
      & 0.180
      & 0.390
      & 0.703
      & 0.568
      & 0.691
      & \textbf{0.230}
      & 0.370 \\

    \midrule
    \multicolumn{13}{l}{\textbf{Image aesthetic assessment models}} \\
    NIMA~\cite{talebi2018nima}
      & \textit{TIP 2018} & 0.054
      & 0.295
      & 0.192
      & 0.555
      & 0.060
      & 0.220
      & 0.357
      & 0.255
      & 0.586
      & 0.080
      & 0.210 \\
    MUSIQ~\cite{ke2021musiq}
      & \textit{ICCV 2021} & 0.027
      & 0.107
      & 0.026
      & 0.506
      & 0.030
      & 0.090
      & 0.230
      & 0.156
      & 0.545
      & 0.080
      & 0.190 \\
    VILA~\cite{ke2023vila}
      & \textit{CVPR 2023} & 0.105
      & 0.409
      & 0.275
      & 0.592
      & \textbf{0.120}
      & \textbf{0.280}
      & 0.355
      & 0.229
      & 0.574
      & \textbf{0.130}
      & 0.290 \\
    AesCLIP~\cite{sheng2023aesclip}
      & \textit{ACM MM 2023} & 0.150
      & 0.477
      & 0.354
      & 0.613
      & 0.030
      & 0.220
      & 0.497
      & 0.351
      & 0.614
      & 0.070
      & 0.260 \\
    EAT~\cite{he2023eat}
      & \textit{ACM MM 2023} & 0.087
      & 0.317
      & 0.211
      & 0.560
      & 0.070
      & 0.160
      & 0.373
      & 0.255
      & 0.585
      & 0.110
      & 0.260 \\
    Q-Align~\cite{wu2024qalign}
      & \textit{ICML 2024} & 8.205
      & 0.457
      & 0.329
      & 0.608
      & 0.100
      & \textbf{0.280}
      & 0.477
      & 0.339
      & 0.610
      & 0.110
      & \textbf{0.300} \\
    Charm~\cite{behrad2025charm}
      & \textit{CVPR 2025} & 0.303
      & 0.401
      & 0.239
      & 0.577
      & 0.050
      & 0.200
      & 0.407
      & 0.229
      & 0.571
      & 0.070
      & 0.220 \\
    HumanAesExpert~\cite{liao2025humanaesexpert}
      & \textit{arXiv 2025} & 8.075
      & -0.014
      & -0.008
      & 0.486
      & 0.050
      & 0.120
      & 0.038
      & 0.025
      & 0.497
      & 0.040
      & 0.130 \\
    ArtiMuse~\cite{cao2026artimuse}
      & \textit{CVPR 2026} & 7.944
      & 0.354
      & 0.223
      & 0.569
      & 0.030
      & 0.160
      & 0.467
      & 0.301
      & 0.595
      & 0.120
      & 0.290 \\
    FGAesQ~\cite{fgaesq2026}
      & \textit{CVPR 2026} & 0.150
      & \textbf{0.522}
      & \textbf{0.368}
      & \textbf{0.619}
      & 0.050
      & 0.140
      & \textbf{0.547}
      & \textbf{0.373}
      & \textbf{0.617}
      & 0.050
      & 0.180 \\
    UniPercept~\cite{cao2026unipercept}
      & \textit{ICML 2026} & 7.944
      & 0.371
      & 0.268
      & 0.558
      & 0.040
      & 0.190
      & 0.453
      & 0.279
      & 0.571
      & 0.070
      & 0.230 \\

    \midrule
    \multicolumn{13}{l}{\textbf{Composition and crop scoring models}} \\
    GAIC evaluator~\cite{zeng2020gaicd}
      & \textit{TPAMI 2020} & 0.003
      & \textbf{0.536}
      & \textbf{0.410}
      & \textbf{0.632}
      & 0.120
      & 0.310
      & \textbf{0.591}
      & \textbf{0.446}
      & \textbf{0.647}
      & 0.080
      & \textbf{0.310} \\
    SAMP-Net~\cite{zhang2021sampnet}
      & \textit{BMVC 2021} & 0.046
      & 0.179
      & 0.126
      & 0.538
      & 0.060
      & 0.150
      & 0.183
      & 0.134
      & 0.543
      & 0.080
      & 0.230 \\
    PhotoFramer~\cite{you2025photoframer}
      & \textit{CVPR 2026} & 8.292
      & 0.468
      & 0.345
      & 0.499
      & 0.080
      & 0.250
      & 0.566
      & 0.408
      & 0.534
      & \textbf{0.130}
      & 0.300 \\
    PortraitCraft~\cite{sha2026portraitcraft}
      & \textit{arXiv 2026} & 4.438
      & 0.453
      & 0.366
      & 0.383
      & \textbf{0.150}
      & \textbf{0.330}
      & 0.532
      & 0.426
      & 0.418
      & 0.060
      & 0.260 \\

    \midrule
    \multicolumn{13}{l}{\textbf{Human preference-aligned reward models}} \\
    ImageReward~\cite{xu2023imagereward}
      & \textit{NeurIPS 2023} & 0.447
      & 0.290
      & 0.208
      & 0.560
      & 0.030
      & \textbf{0.230}
      & 0.371
      & 0.242
      & 0.578
      & 0.050
      & 0.190 \\
    HPSv3~\cite{ma2025hpsv3}
      & \textit{ICCV 2025} & 8.292
      & 0.316
      & 0.184
      & 0.555
      & 0.060
      & 0.110
      & 0.383
      & 0.269
      & 0.586
      & 0.060
      & 0.220 \\
    EditReward~\cite{editreward2026}
      & \textit{ICLR 2026} & 8.297
      & \textbf{0.400}
      & \textbf{0.265}
      & \textbf{0.581}
      & \textbf{0.070}
      & 0.220
      & \textbf{0.529}
      & \textbf{0.380}
      & \textbf{0.625}
      & \textbf{0.100}
      & \textbf{0.260} \\

    \midrule
    \multicolumn{13}{l}{\textbf{Fine-tuned on \cropbench}} \\
    GAIC evaluator~\cite{zeng2020gaicd}
      & \textit{TPAMI 2020} & 0.003
      & \textbf{0.709}
      & \textbf{0.628}
      & \textbf{0.711}
      & \textbf{0.160}
      & \textbf{0.400}
      & \textbf{0.723}
      & \textbf{0.617}
      & \textbf{0.708}
      & \textbf{0.200}
      & \textbf{0.400} \\
    Charm~\cite{behrad2025charm}
      & \textit{CVPR 2025} & 0.303
      & 0.506
      & 0.416
      & 0.623
      & 0.130
      & 0.280
      & 0.575
      & 0.449
      & 0.639
      & 0.110
      & 0.270 \\
    FGAesQ~\cite{fgaesq2026}
      & \textit{CVPR 2026} & 0.150
      & 0.562
      & 0.432
      & 0.633
      & 0.130
      & 0.250
      & 0.606
      & 0.442
      & 0.636
      & 0.090
      & 0.240 \\

    \midrule
    \textbf{CropReward}
      & -- & 4.604
      & \textbf{0.833}
      & \textbf{0.763}
      & \textbf{0.761}
      & \textbf{0.300}
      & \textbf{0.600}
      & \textbf{0.865}
      & \textbf{0.787}
      & \textbf{0.777}
      & \textbf{0.320}
      & \textbf{0.590} \\
    \bottomrule
  \end{tabular}%
  }
  \vspace{2pt}
\end{table*}

\textbf{Evaluation protocols.} On \cropbench, we compute metrics for each set of 15 candidates with the original image included, and average the results over images within each domain. We measure agreement between predicted scores and BT preference scores using the Pearson linear correlation coefficient (PLCC) and Spearman rank correlation coefficient (SRCC). $\mathrm{ACC}_{\mathrm{pair}}$ measures the accuracy of predicting a preference for A, a preference for B, or equivalence (E) over the 105 annotated pairs of each image. $\mathrm{ACC}_{\mathrm{top1}}$ and $\mathrm{ACC}_{\mathrm{top3}}$ measure how often the candidate with the highest predicted score is among the top one or top three candidates ranked by their BT preference scores, respectively. On FGAesthetics-Crop, $s$-ACC measures the accuracy of selecting the most preferred crop in each series, while $s$-SRCC measures the rank correlation within each series.

\textbf{Implementation details.} We train \cropreward\ for three epochs using AdamW, with a learning rate of $2.5{\times}10^{-5}$ and a weight decay of 0.01. The learning rate is linearly warmed up over the first 3\% of training steps and then decayed to zero with a cosine schedule. The original image and candidate crop are resized independently while preserving their aspect ratios, with each image containing at most $512{\times}512$ pixels. LoRA adapters with rank 16, scaling factor 16, and dropout 0.05 are applied to the attention and feed-forward projection layers of the language model. The scale $S$ in the NPD loss (Eq.~\eqref{eq:loss}) is estimated from the training split and fixed at 10.38. For \croprl, we first fine-tune Qwen3.5-4B on the full \cropdata\ for two epochs using LoRA with rank 16. GRPO training then runs for one epoch with a group size of $G=4$, a KL coefficient of 0.01, a format bonus of $\lambda_f=0.05$, and a sampling temperature of 1.0. All training runs use two NVIDIA RTX PRO 6000 Blackwell GPUs.

\begin{table}[!t]
  \centering
\scriptsize 
\caption{Generalization to two external benchmarks,
  GAICD-v2~\cite{zeng2020gaicd} and FGAesthetics-Crop~\cite{fgaesq2026},
  with all models applied without adaptation.}
  \label{tab:metric-generalization}
  \setlength{\tabcolsep}{2.6pt}
  \renewcommand{\arraystretch}{1.10}
  \resizebox{\columnwidth}{!}{%
  \begin{tabular}{l@{\hspace{6pt}}cc@{\hspace{6pt}}cccc}
    \toprule
    \multirow{2}{*}{\textbf{Method}}
      & \multicolumn{2}{c}{\textbf{GAICD-v2}~\cite{zeng2020gaicd}}
      & \multicolumn{4}{c}{\textbf{FGAesthetics-Crop}~\cite{fgaesq2026}} \\
    \cmidrule(lr){2-3}\cmidrule(lr){4-7}
      & $\mathrm{PLCC}$ & $\mathrm{SRCC}$
      & $\mathrm{ACC}$ & $\mathrm{F1}$
      & $s\mbox{-}\mathrm{ACC}$ & $s\mbox{-}\mathrm{SRCC}$ \\
    \midrule
    NIMA~\cite{talebi2018nima}            & 0.232 & 0.210 & 0.658 & 0.653 & 0.303 & 0.323 \\
    MUSIQ~\cite{ke2021musiq}           & 0.184 & 0.164 & \underline{0.753} & \underline{0.753} & \underline{0.406} & \underline{0.499} \\
    VILA~\cite{ke2023vila}            & 0.351 & 0.332 & 0.733 & 0.733 & \underline{0.406} & 0.458 \\
    AesCLIP~\cite{sheng2023aesclip}         & 0.295 & 0.273 & 0.679 & 0.674 & 0.325 & 0.374 \\
    ArtiMuse~\cite{cao2026artimuse}        & \underline{0.413} & \underline{0.386} & 0.682 & 0.682 & 0.354 & 0.387 \\
    FGAesQ~\cite{fgaesq2026}          & 0.294 & 0.275 & \indom{0.775} & \indom{0.775} & \indom{0.473} & \indom{0.555} \\
    GAIC evaluator~\cite{zeng2020gaicd}  & \indom{0.873} & \indom{0.849} & 0.629 & 0.628 & 0.245 & 0.307 \\
    ImageReward~\cite{xu2023imagereward}     & 0.238 & 0.223 & 0.561 & 0.557 & 0.240 & 0.145 \\
    \midrule
    \textbf{CropReward} & \textbf{0.766} & \textbf{0.716} & \textbf{0.822} & \textbf{0.822} & \textbf{0.526} & \textbf{0.662} \\
    \bottomrule
  \end{tabular}%
  }

  \par\vspace{3pt}
  \begin{minipage}{\columnwidth}
  \scriptsize
  \raggedright
  \textit{Note:} PLCC and SRCC are computed
  within each image and macro-averaged; ACC and F1 are pair-level,
  whereas the $s$-prefixed metrics are series-level. Grey cells mark
  \emph{in-domain} results. Bold and underline indicate the best and
  second-best generalization results in each column.
  \end{minipage}
  \vspace{-10pt}
\end{table}

\subsection{Alignment with Human Preferences}\label{sec:exp-agreement}

Table~\ref{tab:crop-metric-benchmark} reports agreement with human cropping preferences on \cropbench. The comparison includes geometry-based metrics, off-the-shelf models for IAA, ICA and human preference modeling, three existing evaluators fine-tuned on \cropbench, and \cropreward. We have the following observations.

\begin{figure*}[!t]
\centering
\includegraphics[width=\textwidth]{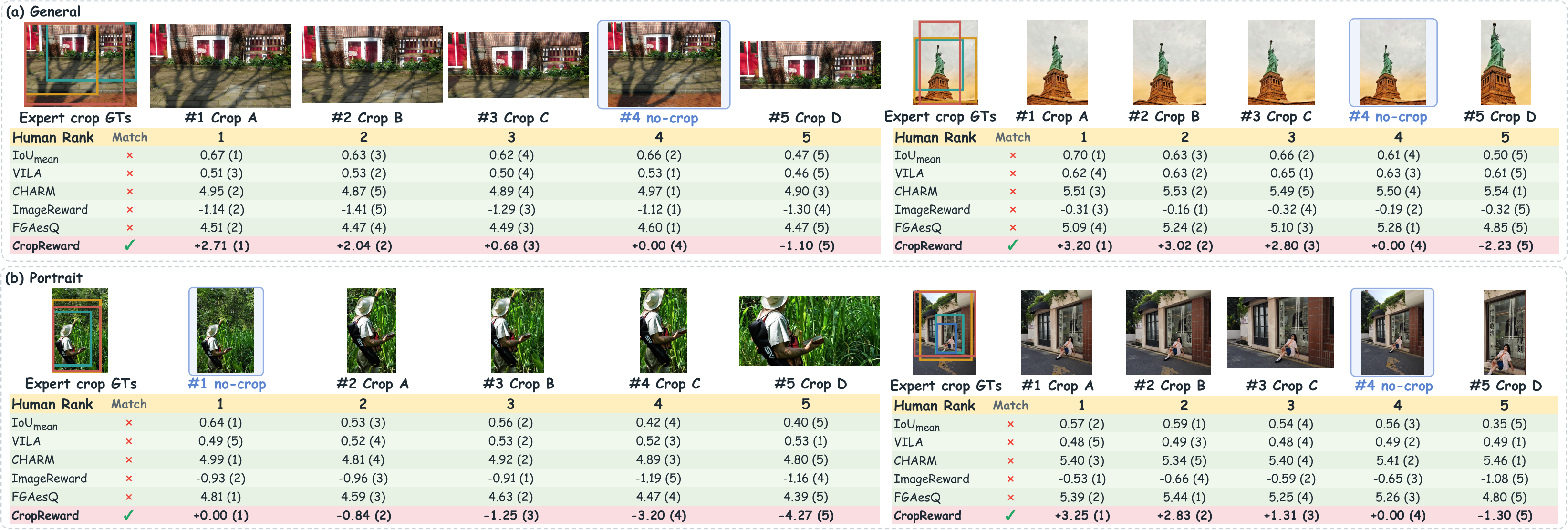}
\caption{Qualitative comparison of predicted and human preference
rankings on \cropbench\ test images from (a) the portrait domain and
(b) the general domain.}
\label{fig:qualitative}
\end{figure*}

\textbf{Expert crop boxes remain valuable references for AIC evaluation.} IoU$_{\mathrm{mean}}$ outperforms all evaluated off-the-shelf models on every metric for both image categories, demonstrating the value of high-quality expert crop boxes for evaluating cropping results. Among the three IoU variants, it also gives the best or joint-best results. Averaging IoU over all three expert crop boxes reduces dependence on any single reference. Nevertheless, agreement with reference boxes does not fully reflect human preferences. An image can have multiple well-composed crops, and a crop that differs from the reference boxes may still be preferred to one with higher IoU. These IoU scores also require expert-annotated reference boxes, limiting their use as rewards on unlabeled images.

\textbf{Off-the-shelf models perform worse than geometry-
based metric.} FGAesQ~\cite{fgaesq2026} achieves the highest PLCC, SRCC, and pairwise accuracy among the IAA models. Its training on relative rankings of visually similar images, including crops, may help it distinguish subtle aesthetic differences between candidates. The GAIC evaluator~\cite{zeng2020gaicd} performs best on these three metrics among all off-the-shelf composition and crop scoring models. It was trained on mean opinion scores for predefined crop candidates. This result suggests that crop-specific supervision is useful for learning human cropping preferences. Among the reward models, EditReward~\cite{editreward2026} performs best on most metrics. Unlike ImageReward and HPSv3, EditReward takes both the original image and the edited result as input, which may help explain its stronger agreement with human cropping preferences than the other two reward models. Across all off-the-shelf models, however, top-1 accuracy reaches at most 0.150 on general images and 0.130 on portraits, showing that they rarely select the candidate most preferred by annotators.

\textbf{\cropreward\ aligns most closely with human preferences.} To compare models trained on the same preference annotations, we also fine-tune the GAIC evaluator, Charm, and FGAesQ on the \cropbench\ training split following their official implementations. As shown in Table~\ref{tab:crop-metric-benchmark}, fine-tuning improves all three models, but \cropreward\ still outperforms all evaluated criteria on every metric for both image categories. Without using any expert-annotated crop boxes as references, \cropreward\ achieves state-of-the-art performance, with SRCC of 0.763 and 0.787 on general and portrait images, respectively, exceeding IoU$_{\mathrm{mean}}$ by 0.140 (22.47\%) and 0.130 (19.79\%). It also exceeds the strongest fine-tuned baseline, the GAIC evaluator, by 0.135 (21.50\%) and 0.170 (27.55\%) in SRCC, respectively. Pairwise accuracy reaches 0.761 and 0.777, while top-3 accuracy reaches 0.600 and 0.590. We examine the effects of the model input and training objective through ablation studies in Sec.~\ref{sec:exp-ablation}.

\textbf{Qualitative analysis.} Fig.~\ref{fig:qualitative} compares existing AIC evaluation criteria with \cropreward\ on four test examples. Existing criteria can disagree with human preferences both when comparing crops with the original image and when ranking alternative crops. In the first general scene example, FGAesQ ranks the original image above crops A, B, and C, although annotators prefer all three crops to the original. Selecting its highest-scoring candidate would retain the original image rather than a crop preferred by annotators. In the second general scene example, Charm assigns its highest score to crop D, which annotators rank below the original image and all other displayed crops. In contrast, \cropreward\ aligns more closely with human preferences than the compared evaluation criteria in these examples. In the first portrait scene example, \cropreward\ assigns negative scores to all four crops, correctly reflecting the annotators' preference for the original image. Even with three expert reference boxes, IoU$_{\mathrm{mean}}$ can misrank visually similar candidates: in the second portrait scene example, it places crop B above A, contrary to human preferences. Higher overlap with expert boxes therefore does not necessarily indicate a more preferred crop.

\begin{table*}[!t]
\centering
\caption{Ablation on the training objective and the input protocol of
\cropreward. All variants use Qwen3.5-4B and are trained for three epochs
under identical settings. \emph{Original} indicates whether
the original image accompanies the candidate crop as input. }
\label{tab:loss_ablation}
\scriptsize 
\setlength{\tabcolsep}{3.2pt}
\renewcommand{\arraystretch}{1.08}
\begin{tabular}{cccc ccccc ccccc}
\toprule
\multirow{2}{*}{Original} &
\multirow{2}{*}{$\mathcal L_{\mathrm{score}}$} &
\multirow{2}{*}{$\mathcal L_{\mathrm{rank}}$} &
\multirow{2}{*}{$\mathcal L_{\mathrm{NPD}}$} &
\multicolumn{5}{c}{General} &
\multicolumn{5}{c}{Portrait} \\
\cmidrule(lr){5-9}\cmidrule(lr){10-14}
& & & & PLCC & SRCC & $\mathrm{ACC}_{\mathrm{pair}}$ &
$\mathrm{ACC}_{\mathrm{top1}}$ & $\mathrm{ACC}_{\mathrm{top3}}$ &
PLCC & SRCC & $\mathrm{ACC}_{\mathrm{pair}}$ &
$\mathrm{ACC}_{\mathrm{top1}}$ & $\mathrm{ACC}_{\mathrm{top3}}$ \\
\midrule
\ding{51} & \ding{51} & &
& 0.824 & 0.754 & 0.755 & 0.260 & 0.560
& 0.854 & 0.758 & 0.764 & 0.250 & \textbf{0.600} \\
\ding{51} & & \ding{51} &
& 0.832 & 0.757 & 0.755 & 0.280 & 0.570
& 0.851 & 0.777 & 0.770 & 0.260 & 0.570 \\
& & & \ding{51}
& 0.715 & 0.607 & 0.706 & 0.180 & 0.450
& 0.778 & 0.624 & 0.710 & 0.130 & 0.450 \\
\ding{51} & & & \ding{51}
& \textbf{0.833} & \textbf{0.763} & \textbf{0.761} & \textbf{0.300} & \textbf{0.600}
& \textbf{0.865} & \textbf{0.787} & \textbf{0.777} & \textbf{0.320} & 0.590 \\
\bottomrule
\end{tabular}
% Keep Tables IV and V together to preserve their order before Fig. 10.
\par\vspace{\floatsep}

\centering
\caption{Comparison of cropping models on CropBench, FLMS, and FCDB.
Best results within each benchmark are bolded. \cropdata\ requires no additional human annotation.}
\label{tab:crop-metric-overall}

\scriptsize
\setlength{\tabcolsep}{1pt}
\renewcommand{\arraystretch}{1.0}

\resizebox{0.97\textwidth}{!}{%
\begin{tabular}{@{}llc@{\hspace{5pt}}c@{\hspace{5pt}}ccccccccc@{}}
\toprule
\multirow{2}{*}{\textbf{Method}}
& \multirow{2}{*}{\textbf{Training data}}
& \multicolumn{10}{c}{\textbf{Existing metrics}}
& \multicolumn{1}{c}{\textbf{Proposed metric}} \\
\cmidrule(lr){3-12}
\cmidrule(l){13-13}
& & \textbf{IoU} & \textbf{Disp}
& \shortstack{\textbf{VILA}\\\cite{ke2023vila}}
& \shortstack{\textbf{AesCLIP}\\\cite{sheng2023aesclip}}
& \shortstack{\textbf{Charm}\\\cite{behrad2025charm}}
& \shortstack{\textbf{ArtiMuse}\\\cite{cao2026artimuse}}
& \shortstack{\textbf{FGAesQ}\\\cite{fgaesq2026}}
& \shortstack{\textbf{PhotoFramer}\\\cite{you2025photoframer}}
& \shortstack{\textbf{EditReward}\\\cite{editreward2026}}
& \textbf{\emph{Overall}}
& \textbf{CropReward} \\
\midrule

% ======================== CropBench ========================

\multicolumn{13}{@{}l}{\textbf{\textit{CropBench}}} \\[1pt]
A2-RL~\cite{li2018a2rl}
    &   AVA~\cite{murray2012ava}
    &   0.623  &  0.109  &   0.457   &   5.025   &   5.095   &   53.65
    &   4.703   &   2.977   &   0.540   &   0.5494   &   1.28 \\
CACNet~\cite{hong2021cacnet}
    &   FCDB + KU-PCP~\cite{chen2017quantitative,lee2018kupcp}
    &   0.641  &  0.098  &   0.457   &   5.013   &   5.074   &   53.77
    &   4.664   &   2.962   &   \textbf{0.577}   &   0.5524   &   1.34 \\
CGS~\cite{li2020composing}
    &   GAICD-v1~\cite{zeng2019reliable}
    &   0.622  &  0.101  &   0.453   &   5.016   &   5.049   &   52.63
    &   4.639   &   2.989   &   0.516   &   0.5468   &   1.12 \\
GAIC~\cite{zeng2020gaicd}
    &   GAICD-v2~\cite{zeng2020gaicd}
    &   0.639  &  0.097  &   0.454   &   5.039   &   5.063   &   53.18
    &   4.665   &   2.992   &   0.559   &   0.5519   &   1.42 \\
HCIC~\cite{zhang2022hcic}
    &   CPC~\cite{wei2018good}
    &   0.655  &  0.093  &   0.459   &   5.014   &   5.061   &   53.58
    &   4.675   &   2.995   &   0.548   &   0.5547   &   1.81 \\
InstructCrop~\cite{sheng2025instructcrop}
    &   FCDB~\cite{chen2017quantitative}
    &   0.634  &  0.096  &   0.453   &   5.005   &   5.057   &   53.27
    &   4.643   &   2.930   &   0.561   &   0.5489   &   1.09 \\
S2CNet~\cite{s2cnet2023}
    &   GAICD-v1~\cite{zeng2019reliable}
    &   0.632  &  0.097  &   0.454   &   5.041   &   5.053   &   53.00
    &   4.660   &   2.985   &   0.557   &   0.5503   &   1.43 \\
Venus~\cite{venus2026}
    &   CPC+GAICD-v2+FCDB~\cite{wei2018good,zeng2020gaicd,chen2017quantitative}
    &   0.660  &  0.091  &   0.458   &   5.019   &   \textbf{5.101}
    &   \textbf{53.84}   &   4.670   &   2.981   &   0.383
    &   0.5516   &   1.69 \\
GPT-5.5~\cite{openai2026gpt55}
    &   \textit{--}
    &   0.665  &  0.089  &   0.456   &   5.027   &   5.059   &   53.31
    &   4.664   &   2.999   &   0.535   &   0.5554   &   1.83 \\
\noalign{\vskip 3pt}
\hdashline
\noalign{\vskip 3pt}
\croprlsft
    &   \multirow{2}{*}{\begin{tabular}[c]{@{}l@{}}\cropdata\\[-1pt](no additional human annotation)\end{tabular}}
    &   0.675  &  0.085  &   0.459   &   5.029   &   5.060   &   53.63
    &   4.702   &   3.004   &   0.545   &   0.5585   &   2.55 \\
\textbf{\croprl}
    &  
    &   \textbf{0.698}  &  \textbf{0.076}  &   \textbf{0.461}   &   \textbf{5.047}
    &   5.048   &   53.76   &   \textbf{4.733}   &   \textbf{3.043}
    &   0.531   &   \textbf{0.5635}   &   \textbf{3.05} \\

\midrule

% =========================== FLMS ==========================
\multicolumn{13}{@{}l}{\textbf{\textit{FLMS}}~\cite{fang2014automatic}} \\[1pt]
A2-RL
    &   AVA
    &   0.821  &  0.045  &   0.472   &   5.108   &   5.148   &   59.04
    &   5.083   &   3.109   &   0.602   &   0.5978   &   1.77 \\
CACNet
    &   FCDB + KU-PCP
    &   0.841  &  0.037  &   0.471   &   5.098   &   5.145   &   59.10
    &   5.093   &   3.166   &   0.636   &   0.6033   &   1.70 \\
CGS
    &   GAICD-v1
    &   0.797  &  0.047  &   0.463   &   5.061   &   5.136   &   58.29
    &   5.062   &   3.045   &   0.608   &   0.5904   &   1.35 \\
GAIC
    &   GAICD-v2
    &   0.821  &  0.043  &   0.469   &   5.109   &   5.142   &   58.79
    &   5.102   &   \textbf{3.281}   &   0.620   &   0.6030   &   2.06 \\
HCIC
    &   CPC
    &   0.840  &  0.037  &   0.471   &   5.096   &   5.145   &   59.23
    &   5.098   &   3.094   &   0.628   &   0.6012   &   2.07 \\
InstructCrop
    &   FCDB
    &   0.827  &  0.040  &   0.469   &   5.091   &   5.141   &   59.08
    &   5.089   &   3.144   &   0.636   &   0.6004   &   1.40 \\
S2CNet
    &   GAICD-v1
    &   0.810  &  0.045  &   0.468   &   5.110   &   5.145   &   58.58
    &   5.095   &   3.203   &   0.621   &   0.5989   &   1.89 \\
Venus
    &   CPC + GAICD-v2 + FCDB
    &   \textbf{0.864}  &  \textbf{0.030}  &   \textbf{0.474}   &   5.096
    &   \textbf{5.152}   &   \textbf{59.42}
    &   5.094   &   3.090   &   0.567   &   0.6037   &   1.65 \\
GPT-5.5
    &   \textit{--}
    &   0.841  &  0.037  &   0.472   &   5.122   &   5.143   &   59.15
    &   5.107   &   3.166   &   \textbf{0.648}
    &   \textbf{0.6042}   &   2.30 \\
\noalign{\vskip 3pt}
\hdashline
\noalign{\vskip 3pt}
\croprlsft
    &   \multirow{2}{*}{\begin{tabular}[c]{@{}l@{}}\cropdata\\[-1pt](no additional human annotation)\end{tabular}}
    &   0.841  &  0.037  &   0.473   &   5.118   &   5.142   &   59.20
    &   \textbf{5.118}   &   3.156   &   0.623   &   0.6036   &   2.52 \\
\textbf{\croprl}
    &  
    &   0.841  &  0.037  &   0.473   &   \textbf{5.127}   &   5.143   &   59.32
    &   5.112   &   3.167   &   0.623
    &   \textbf{0.6042}   &   \textbf{2.82} \\

\midrule

% =========================== FCDB ==========================
\multicolumn{13}{@{}l}{\textbf{\textit{FCDB}}~\cite{chen2017quantitative}} \\[1pt]
A2-RL
    &   AVA
    &   0.665  &  0.089  &   \textbf{0.432}   &   \textbf{4.851}
    &   \textbf{4.968}   &   51.10
    &   4.868   &   2.908   &   0.368   &   0.5427   &   1.72 \\
CACNet
    &   FCDB + KU-PCP
    &   0.696  &  0.075  &   0.427   &   4.795   &   4.966   &   \textbf{51.27}
    &   4.829   &   2.908   &   \textbf{0.383}   &   0.5465   &   1.63 \\
CGS
    &   GAICD-v1
    &   0.657  &  0.086  &   0.414   &   4.717   &   4.890   &   50.34
    &   4.782   &   2.833   &   0.315   &   0.5320   &   1.22 \\
GAIC
    &   GAICD-v2
    &   0.678  &  0.082  &   0.426   &   4.829   &   4.946   &   50.86
    &   4.826   &   \textbf{3.024}   &   0.372   &   0.5462   &   1.88 \\
HCIC
    &   CPC
    &   0.700  &  0.075  &   0.428   &   4.799   &   4.955   &   51.13
    &   4.874   &   2.931   &   0.369   &   0.5475   &   1.93 \\
InstructCrop
    &   FCDB
    &   0.687  &  0.077  &   0.423   &   4.778   &   4.947   &   51.03
    &   4.807   &   2.854   &   0.377   &   0.5422   &   1.48 \\
S2CNet
    &   GAICD-v1
    &   0.675  &  0.083  &   0.421   &   4.811   &   4.930   &   50.61
    &   4.813   &   2.962   &   0.362   &   0.5423   &   1.73 \\
Venus
    &   CPC + GAICD-v2 + FCDB
    &   0.682  &  0.080  &   0.424   &   4.805   &   4.936   &   50.54
    &   4.845   &   2.990   &   0.325   &   0.5438   &   1.88 \\
GPT-5.5
    &   \textit{--}
    &   0.697  &  0.076  &   0.427   &   4.823   &   4.948   &   51.06
    &   4.864   &   2.937   &   0.366   &   0.5472   &   2.14 \\
\noalign{\vskip 3pt}
\hdashline
\noalign{\vskip 3pt}
\croprlsft
    &   \multirow{2}{*}{\begin{tabular}[c]{@{}l@{}}\cropdata\\[-1pt](no additional human annotation)\end{tabular}}
    &   0.705  &  0.074  &   0.431   &   4.846   &   4.957   &   50.96
    &   4.867   &   2.972   &   0.370   &   0.5501   &   2.39 \\
\textbf{\croprl}
    &  
    &   \textbf{0.706}  &  \textbf{0.073}  &   0.429   &   4.846   &   4.945   &   51.09
    &   \textbf{4.875}   &   3.002   &   0.361
    &   \textbf{0.5509}   &   \textbf{2.60} \\
\bottomrule
\end{tabular}%
}
\vspace{-10pt}
\end{table*}

\subsection{Generalization to External Benchmarks}\label{sec:exp-generalization}We evaluate \cropreward\ on GAICD-v2 and FGAesthetics-Crop to examine its generalization beyond \cropbench. The two benchmarks differ from \cropbench\ in their candidate crops and annotation types. GAICD-v2 provides MOS for grid-anchor candidates, whereas FGAesthetics-Crop provides pairwise preferences among visually similar crops derived from GAICD and CPC~\cite{wei2018good}. We use \cropreward\ trained on \cropbench\ without further fine-tuning. Table~\ref{tab:metric-generalization} compares its performance with representative existing models, including models trained on the corresponding benchmark. On GAICD-v2, \cropreward\ achieves PLCC of 0.766 and SRCC of 0.716, outperforming all compared out-of-domain models. The strongest of these baselines, ArtiMuse, achieves 0.413 and 0.386, respectively. On FGAesthetics-Crop, \cropreward\ achieves pair-level ACC and F1 of 0.822 and series-level $s$-ACC and $s$-SRCC of 0.526 and 0.662. It even \textbf{surpasses the in-domain FGAesQ} on all four metrics without fine-tuning on FGAesthetics. These results indicate that \cropreward, trained on \cropbench, aligns with both human quality ratings and fine-grained crop preferences on the two external benchmarks.

\begin{figure*}[!t]
    \centering
    \includegraphics[width=\textwidth]{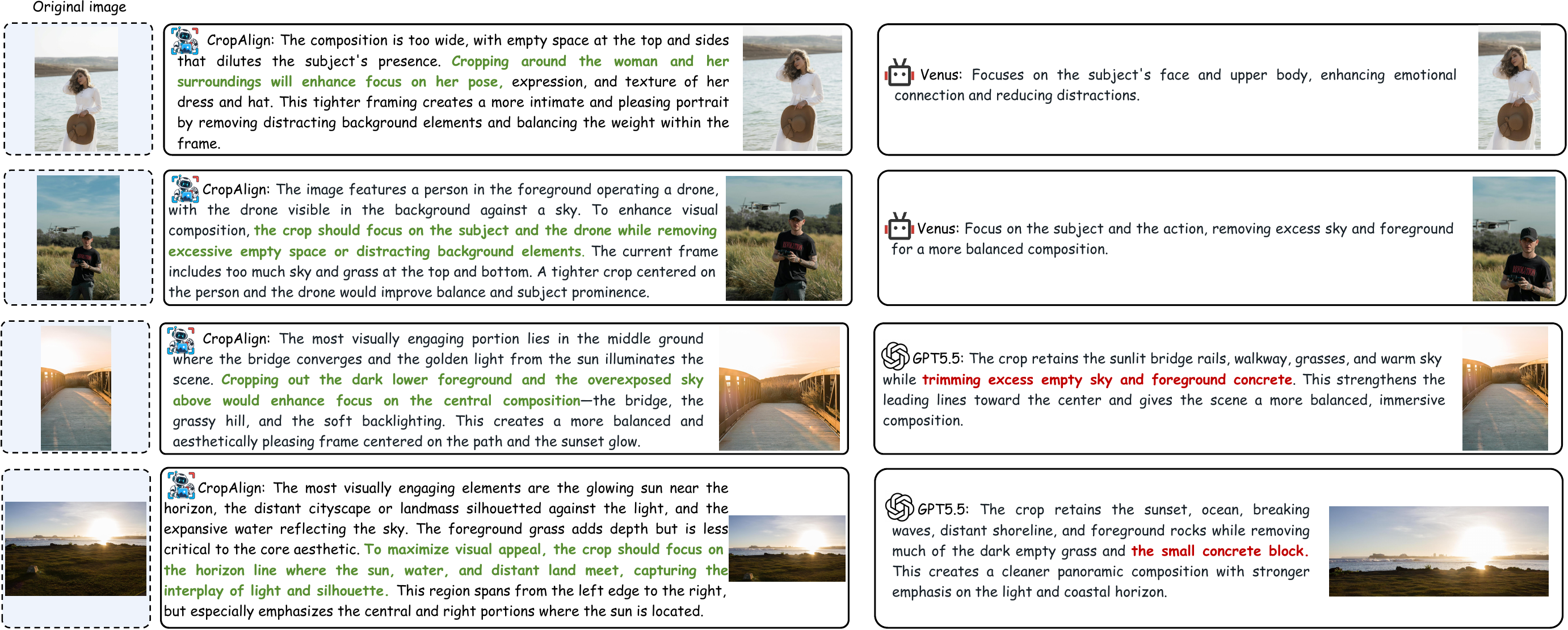}
\caption{Comparison of explanations for cropping decisions. Green
and red highlight selected crop-related phrases from \croprl\ and
GPT-5.5, respectively.}
    \label{fig:interpretability}
    \vspace{-10pt}
\end{figure*}

\begin{table}[t]
  \centering
  \caption{User study comparing \croprl\ with six competing methods.}
  \label{tab:user-study}
  \setlength{\tabcolsep}{4pt}
  \renewcommand{\arraystretch}{1.10}
  \begin{tabular}{lccc}
    \toprule
    \textbf{Comparison} & \textbf{Win (\%)} & \textbf{Tie (\%)} & \textbf{Lose (\%)} \\
    \midrule
    \textit{vs.} GPT-5.5~\cite{openai2026gpt55}        & 73.2 & 1.4 & 25.4 \\
    \textit{vs.} Venus~\cite{venus2026} & 74.2 & 0.6 & 25.2 \\
    \textit{vs.} HCIC~\cite{zhang2022hcic}           & 75.6 & 0.6 & 23.8 \\
    \textit{vs.} GAIC~\cite{zeng2020gaicd}           & 76.2 & 0.6 & 23.2 \\
    \textit{vs.} CACNet~\cite{hong2021cacnet}         & 76.4 & 0.2 & 23.4 \\
    \textit{vs.} InstructCrop~\cite{sheng2025instructcrop}   & 78.6 & 0.2 & 21.2 \\
    \midrule
    \textbf{Overall}                 & \textbf{75.7} & \textbf{0.6} & \textbf{23.7} \\
    \bottomrule
  \end{tabular}
\end{table}

\subsection{Ablation Studies}\label{sec:exp-ablation}
We examine two design choices of \cropreward\ on the \cropbench\ test split.

\textbf{Training objective.} Table~\ref{tab:loss_ablation} compares the NPD loss in Eq.~\eqref{eq:loss} with pointwise score regression, $\mathcal L_{\mathrm{score}}$, and the pairwise logistic ranking loss, $\mathcal L_{\mathrm{rank}}$, commonly used to train reward models from human preferences~\cite{xu2023imagereward,kirstain2023pick}. All three variants take both the original image and the candidate crop as input. $\mathcal L_{\mathrm{score}}$ supervises individual BT preference scores, whereas our ranking baseline uses preference order and excludes pairs annotated as equivalent. NPD directly matches predicted score differences to BT preference score differences over all candidate pairs. It achieves the best results on nine of the ten metrics across the two domains.

\textbf{Original image input.} It can also be observed from Table~\ref{tab:loss_ablation} that removing the original image from the input while retaining the NPD loss lowers all ten metrics. SRCC drops by 0.156 and 0.163 on general and portrait images, respectively. $\mathrm{ACC}_{\mathrm{top1}}$ falls from 0.300 to 0.180 on general images and from 0.320 to 0.130 on portraits. These results show that original image input helps \cropreward\ predict human cropping preferences.

\subsection{Cropping Performance of \croprl}\label{sec:exp-downstream}

We next evaluate the effectiveness of \cropreward\ in guiding AIC model optimization by comparing the proposed \croprl\ with existing AIC methods.

\textbf{Quantitative comparison.} Table~\ref{tab:crop-metric-overall} compares \croprl\ and its SFT-only variant, \croprlsft, with nine representative AIC methods, including specialized models and MLLMs. We evaluate these methods on the \cropbench\ and FCDB test set and on the full FLMS dataset, following prior AIC studies. On CropBench, IoU and Disp are averaged over the three expert-annotated reference boxes. On FLMS, both metrics are computed against the reference box with the highest IoU, whereas on FCDB, they are computed against the single reference box. Each crop is also scored by seven trained evaluation criteria. For Overall, we normalize Disp as $1-\mathrm{Disp}$ and average it with the normalized values of the other eight metrics; the \cropreward\ score is reported separately and excluded from this average. Even when evaluated using existing criteria, \croprl\ achieves the best \emph{Overall} score on all three benchmarks. Trained on \cropdata, \croprlsft\ is already competitive with the strongest baselines, supporting the use of crops selected by \cropreward\ as training supervision. GRPO further improves the \emph{Overall} score on each benchmark. On \cropbench, IoU also increases from 0.675 to 0.698 after GRPO. The \cropreward\ scores increase as well, consistent with the reward used for optimization.

\textbf{User study.} We further compare \croprl\ with six representative AIC methods through a user study. Each comparison covers 500 images: all 200 \cropbench\ test images, 150 images randomly sampled from FLMS, and 150 from the FCDB test set. Five distinct annotators from a pool of ten judge each pair, with the left/right order randomized. Their votes are aggregated by majority. As reported in Table~\ref{tab:user-study}, \croprl\ wins 75.7\% of the 3,000 pairs. Its win rates range from 73.2\% against GPT-5.5 to 78.6\% against InstructCrop, and all six comparisons are significant at $p<0.001$ under a binomial test. These results show that users prefer crops from \croprl\ over those of the compared specialized models and MLLMs.

\textbf{Interpretability.} When prompted to explain its cropping decisions, \croprl\ can generate a textual rationale followed by a crop box. Fig.~\ref{fig:interpretability} compares its explanations and cropping results with those of Venus and GPT-5.5. In the person-and-drone example, \croprl\ identifies excess sky and grass in the original and reduces these regions while retaining both the person and the drone. In the bridge example, it reduces the sky and lower foreground as described in its rationale, while GPT-5.5 makes smaller adjustments to these regions. These examples illustrate that \croprl\ can provide explanations consistent with its crop boxes without additional supervision from human-written rationales during SFT or GRPO.

\begin{figure}[t]
    \centering
    \includegraphics[width=\columnwidth]{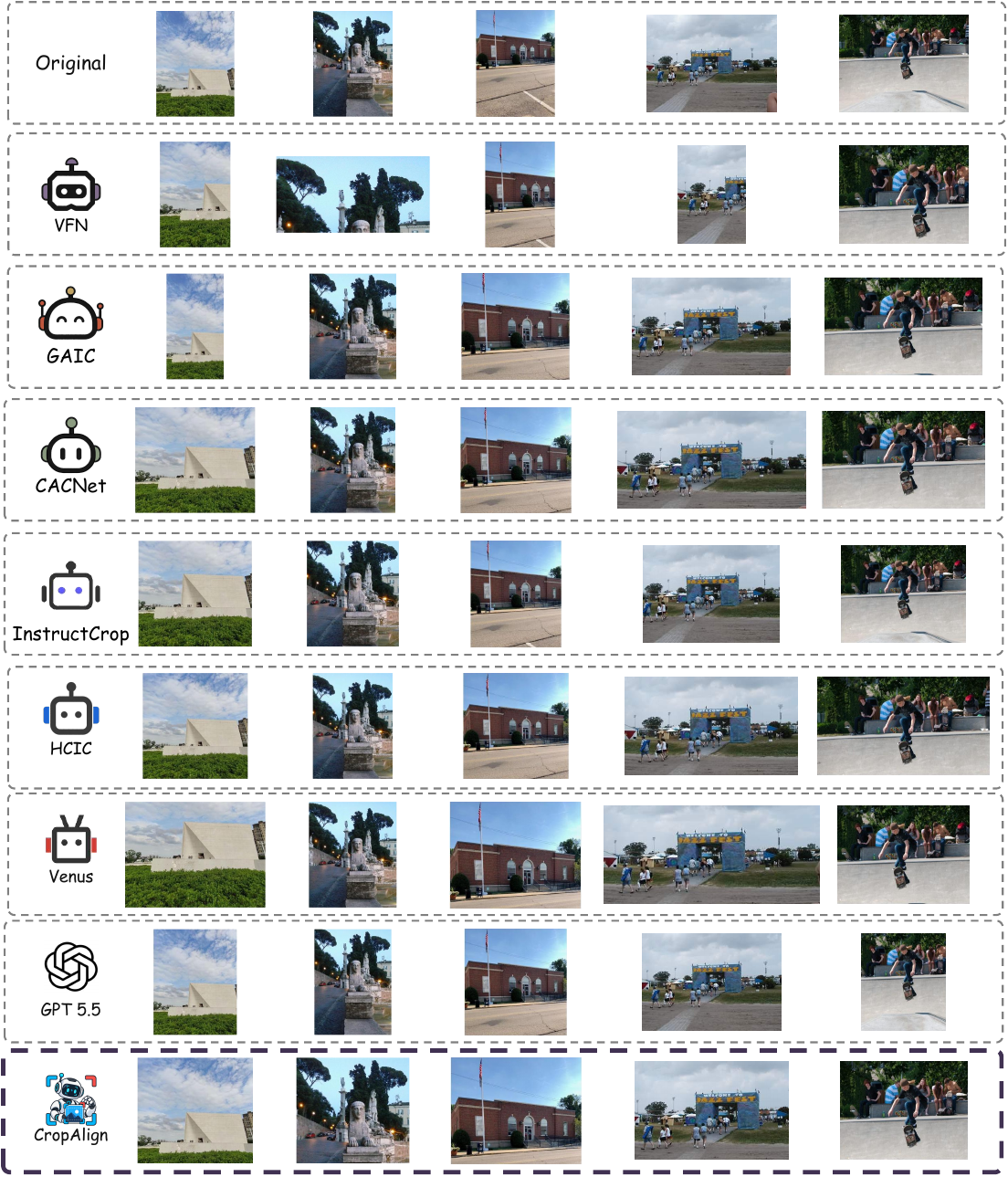}
    \caption{Qualitative comparison of cropping results.  }
    \label{fig:qualitative_crops}
    \vspace{-10pt}
\end{figure}

\textbf{Qualitative comparison.} Fig.~\ref{fig:qualitative_crops} compares \croprl\ with representative AIC methods.In the statue example, \croprl\ preserves the statue and its base, whereas VFN removes the lower part of the subject. Overall, \croprl\ produces better-composed crops in these examples, preserving the main subjects while removing unnecessary surroundings. In the fourth column, \croprl\ crops out more of the empty foreground than GPT-5.5 while retaining the entrance and the people in front of it. This tighter framing places greater emphasis on the entrance and the surrounding activity.

\section{Conclusion}\label{sec:conclusion}

In this paper, we have introduced \textbf{\cropreward}, an MLLM-based reward model for AIC task that assesses improvement or degradation relative to the original image and distinguishes fine-grained preferences among crops. To examine how well existing AIC evaluation criteria reflect human cropping preferences, we contributed \textbf{\cropbench}, which provides pairwise preference annotations among candidate crops and between crops and their original images. Experiments show that \cropreward\ aligns more closely with human preferences than the compared evaluation criteria and generalizes well to external benchmarks. We further employed \cropreward\ to build a new large-scale AIC dataset \textbf{\cropdata}, and further used it to guide the reinforcement learning of \textbf{\croprl}, a new AIC model. This training process does not require any human annotation. \croprl\ delivers the state-of-the-art performance under existing evaluation criteria and is preferred over the specialized models and MLLMs in a user study. It is expected that our work offers a new paradigm for AIC evaluation and optimization based on human preferences.

% \section*{Acknowledgments}
% This work is supported by the National Natural Science Foundation of China under Grants 62471349, 625B2142, 62301378, 62501080, and 62171340; the Fundamental Research Funds for the Central Universities under Grants YJSJ25004 and QTZX25076; and partly by the China Postdoctoral Science Foundation under Grant 2024M762553. The authors thank the annotators from the BRAVE Lab at Xidian University for their valuable contributions to the dataset annotation.

\bibliographystyle{IEEEtran}
\bibliography{references}

@article{zeng2020gaicd,
  title={Grid anchor based image cropping: A new benchmark and an efficient model},
  author={Zeng, Hui and Li, Lida and Cao, Zisheng and Zhang, Lei},
  journal={IEEE Transactions on Pattern Analysis and Machine Intelligence},
  volume={44},
  number={3},
  pages={1304--1319},
  year={2020},
  publisher={IEEE}
}

@inproceedings{yan2013learning,
  title={Learning the change for automatic image cropping},
  author={Yan, Jianzhou and Lin, Stephen and Kang, Sing Bing and Tang, Xiaoou},
  booktitle={Proceedings of the IEEE Conference on Computer Vision and Pattern Recognition},
  pages={971--978},
  year={2013}
}

@inproceedings{fang2014automatic,
  title={Automatic image cropping using visual composition, boundary simplicity and content preservation models},
  author={Fang, Chen and Lin, Zhe and Mech, Radomir and Shen, Xiaohui},
  booktitle={Proceedings of the 22nd ACM International Conference on Multimedia},
  pages={1105--1108},
  year={2014}
}

@inproceedings{chen2017quantitative,
  title={Quantitative analysis of automatic image cropping algorithms: A dataset and comparative study},
  author={Chen, Yi-Ling and Huang, Tzu-Wei and Chang, Kai-Han and Tsai, Yu-Chen and Chen, Hwann-Tzong and Chen, Bing-Yu},
  booktitle={2017 IEEE Winter Conference on Applications of Computer Vision (WACV)},
  pages={226--234},
  year={2017},
  organization={IEEE}
}

@inproceedings{chen2017vfn,
  title={Learning to compose with professional photographs on the web},
  author={Chen, Yi-Ling and Klopp, Jan and Sun, Min and Chien, Shao-Yi and Ma, Kwan-Liu},
  booktitle={Proceedings of the 25th ACM International Conference on Multimedia},
  pages={37--45},
  year={2017}
}

@inproceedings{li2018a2rl,
  title={A2-RL: Aesthetics aware reinforcement learning for image cropping},
  author={Li, Debang and Wu, Huikai and Zhang, Junge and Huang, Kaiqi},
  booktitle={Proceedings of the IEEE Conference on Computer Vision and Pattern Recognition},
  pages={8193--8201},
  year={2018}
}

@inproceedings{wei2018good,
  title={Good view hunting: Learning photo composition from dense view pairs},
  author={Wei, Zijun and Zhang, Jianming and Shen, Xiaohui and Lin, Zhe and Mech, Radomir and Hoai, Minh and Samaras, Dimitris},
  booktitle={Proceedings of the IEEE Conference on Computer Vision and Pattern Recognition},
  pages={5437--5446},
  year={2018}
}

@inproceedings{li2020composing,
  title={Composing good shots by exploiting mutual relations},
  author={Li, Debang and Zhang, Junge and Huang, Kaiqi and Yang, Ming-Hsuan},
  booktitle={Proceedings of the IEEE/CVF Conference on Computer Vision and Pattern Recognition},
  pages={4213--4222},
  year={2020}
}

@inproceedings{hong2021cacnet,
  title={Composing photos like a photographer},
  author={Hong, Chaoyi and Du, Shuaiyuan and Xian, Ke and Lu, Hao and Cao, Zhiguo and Zhong, Weicai},
  booktitle={Proceedings of the IEEE/CVF Conference on Computer Vision and Pattern Recognition},
  pages={7057--7066},
  year={2021}
}

@inproceedings{hong2024gencrop,
  title={Learning subject-aware cropping by outpainting professional photos},
  author={Hong, James and Yuan, Lu and Gharbi, Micha{\"e}l and Fisher, Matthew and Fatahalian, Kayvon},
  booktitle={Proceedings of the AAAI Conference on Artificial Intelligence},
  volume={38},
  number={3},
  pages={2175--2183},
  year={2024}
}

@inproceedings{s2cnet2023,
  title={Spatial-semantic collaborative cropping for user generated content},
  author={Su, Yukun and Cao, Yiwen and Deng, Jingliang and Rao, Fengyun and Wu, Qingyao},
  booktitle={Proceedings of the AAAI Conference on Artificial Intelligence},
  volume={38},
  number={5},
  pages={4988--4997},
  year={2024}
}

@inproceedings{lee2025cropper,
  title={Cropper: Vision-language model for image cropping through in-context learning},
  author={Lee, Seung Hyun and Jiang, Jijun and Xu, Yiran and Li, Zhuofang and Ke, Junjie and Li, Yinxiao and He, Junfeng and Hickson, Steven and Datsenko, Katie and Kim, Sangpil and Yang, Ming-Hsuan and Essa, Irfan and Yang, Feng},
  booktitle={Proceedings of the IEEE/CVF Conference on Computer Vision and Pattern Recognition},
  pages={30010--30019},
  year={2025}
}

@inproceedings{venus2026,
  title={Venus: Benchmarking and Empowering Multimodal Large Language Models for Aesthetic Guidance and Cropping},
  author={Du, Tianxiang and He, Hulingxiao and Peng, Yuxin},
  booktitle={Proceedings of the IEEE/CVF Conference on Computer Vision and Pattern Recognition},
  pages={37766--37776},
  year={2026}
}

@inproceedings{sheng2025instructcrop,
author = {Sheng, Xiangfei and Xie, Pangu and Zou, Weidong and Chen, Pengfei and Zhu, Tong and Li, Leida},
title = {InstructCrop: Teaching Multimodal Large Language Models to Crop Aesthetic Images},
year = {2025},
booktitle = {Proceedings of the 33rd ACM International Conference on Multimedia},
pages = {6830-6839},
}

@ARTICLE{wang2019cropping,
  author={Wang, Wenguan and Shen, Jianbing and Ling, Haibin},
  journal={IEEE Transactions on Pattern Analysis and Machine Intelligence}, 
  title={A Deep Network Solution for Attention and Aesthetics Aware Photo Cropping}, 
  year={2019},
  volume={41},
  number={7},
  pages={1531-1544}
}

@article{xu2023imagereward,
  title={ImageReward: Learning and evaluating human preferences for text-to-image generation},
  author={Xu, Jiazheng and Liu, Xiao and Wu, Yuchen and Tong, Yuxuan and Li, Qinkai and Ding, Ming and Tang, Jie and Dong, Yuxiao},
  journal={Advances in Neural Information Processing Systems},
  volume={36},
  pages={15903--15935},
  year={2023}
}

@article{dong2026crop,
  title={CROP: Expert-Aligned Image Cropping via Compositional Reasoning and Optimizing Preference},
  author={Dong, Zhitong and Li, Chao and Yu, Jie and Chen, Hao},
  journal={arXiv preprint arXiv:2605.12545},
  year={2026}
}

@inproceedings{ma2025hpsv3,
  title={HPSv3: Towards wide-spectrum human preference score},
  author={Ma, Yuhang and Wu, Xiaoshi and Sun, Keqiang and Li, Hongsheng},
  booktitle={Proceedings of the IEEE/CVF International Conference on Computer Vision},
  pages={15086--15095},
  year={2025}
}

@article{editreward2026,
  title={EditReward: A Human-Aligned Reward Model for Instruction-Guided Image Editing},
  author={Wu, Keming and Jiang, Sicong and Ku, Max and Nie, Ping and Liu, Minghao and Chen, Wenhu},
  journal={arXiv preprint arXiv:2509.26346},
  year={2025}
}

@inproceedings{beautygrpo2026,
  title={BeautyGRPO: Aesthetic alignment for face retouching via dynamic path guidance and fine-grained preference modeling},
  author={Yang, Jiachen and Lin, Xianhui and Dong, Yi and Zheng, Zebiao and Liu, Xing and Gu, Hong and Fang, Yanmei},
  booktitle={Proceedings of the IEEE/CVF Conference on Computer Vision and Pattern Recognition},
  pages={25110--25120},
  year={2026}
}

@article{talebi2018nima,
  title={NIMA: Neural image assessment},
  author={Talebi, Hossein and Milanfar, Peyman},
  journal={IEEE transactions on image processing},
  volume={27},
  number={8},
  pages={3998--4011},
  year={2018},
  publisher={IEEE}
}

@inproceedings{ke2023vila,
  title={Vila: Learning image aesthetics from user comments with vision-language pretraining},
  author={Ke, Junjie and Ye, Keren and Yu, Jiahui and Wu, Yonghui and Milanfar, Peyman and Yang, Feng},
  booktitle={Proceedings of the IEEE/CVF Conference on Computer Vision and Pattern Recognition},
  pages={10041--10051},
  year={2023}
}

@article{zhang2021sampnet,
  title={Image Composition Assessment with Saliency-augmented Multi-pattern Pooling},
  author={Zhang, Bo and Niu, Li and Zhang, Liqing},
  journal={arXiv preprint arXiv:2104.03133},
  year={2021}
}

@inproceedings{you2025photoframer,
  title={Photoframer: Multi-modal image composition instruction},
  author={You, Zhiyuan and Wang, Ke and Zhang, He and Cai, Xin and Gu, Jinjin and Xue, Tianfan and Dong, Chao and Zhang, Zhoutong},
  booktitle={Proceedings of the IEEE/CVF Conference on Computer Vision and Pattern Recognition},
  pages={10197--10207},
  year={2026}
}

@inproceedings{zhang2022hcic,
  title={Human-Centric Image Cropping with Partition-Aware and Content-Preserving Features},
  author={Zhang, Bo and Niu, Li and Zhao, Xing and Zhang, Liqing},
  booktitle={European Conference on Computer Vision},
  pages={181--197},
  year={2022},
  organization={Springer},
  doi={10.1007/978-3-031-20071-7_11}
}

@article{lu2019lvrn,
  title={Listwise View Ranking for Image Cropping},
  author={Lu, Weirui and Xing, Xiaofen and Cai, Bolun and Xu, Xiangmin},
  journal={IEEE Access},
  volume={7},
  pages={91904--91911},
  year={2019},
  doi={10.1109/ACCESS.2019.2925430}
}

@inproceedings{zhang2026procrop,
  title={ProCrop: Learning Aesthetic Image Cropping from Professional Compositions},
  author={Zhang, Ke and Ding, Tianyu and Jiang, Jiachen and Chen, Tianyi and Zharkov, Ilya and Patel, Vishal M. and Liang, Luming},
  booktitle={Proceedings of the AAAI Conference on Artificial Intelligence},
  volume={40},
  number={15},
  pages={12600--12608},
  year={2026},
  doi={10.1609/aaai.v40i15.38255}
}

@inproceedings{zeng2019reliable,
  title={Reliable and Efficient Image Cropping: A Grid Anchor Based Approach},
  author={Zeng, Hui and Li, Lida and Cao, Zisheng and Zhang, Lei},
  booktitle={Proceedings of the IEEE/CVF Conference on Computer Vision and Pattern Recognition},
  pages={5949--5957},
  year={2019}
}

@inproceedings{ke2021musiq,
  title={{MUSIQ}: Multi-Scale Image Quality Transformer},
  author={Ke, Junjie and Wang, Qifei and Wang, Yilin and Milanfar, Peyman and Yang, Feng},
  booktitle={Proceedings of the IEEE/CVF International Conference on Computer Vision},
  pages={5148--5157},
  year={2021}
}

@inproceedings{wu2024qalign,
  title={Q-Align: Teaching {LMM}s for Visual Scoring via Discrete Text-Defined Levels},
  author={Wu, Haoning and Zhang, Zicheng and Zhang, Weixia and Chen, Chaofeng and Liao, Liang and Li, Chunyi and Gao, Yixuan and Wang, Annan and Zhang, Erli and Sun, Wenxiu and Yan, Qiong and Min, Xiongkuo and Zhai, Guangtao and Lin, Weisi},
  booktitle={Proceedings of the 41st International Conference on Machine Learning},
  pages={54015--54029},
  year={2024},
  volume={235},
  series={Proceedings of Machine Learning Research},
  publisher={PMLR}
}

@inproceedings{fgaesq2026,
  title={Fine-Grained Image Aesthetic Assessment: Learning Discriminative Scores from Relative Ranks},
  author={Yang, Zhichao and Wang, Jianjie and Zhang, Zhixianhe and Xie, Pangu and Sheng, Xiangfei and Chen, Pengfei and Li, Leida},
  booktitle={Proceedings of the IEEE/CVF Conference on Computer Vision and Pattern Recognition},
  pages={145--155},
  year={2026}
}

@inproceedings{kirstain2023pick,
  title={Pick-a-Pic: An Open Dataset of User Preferences for Text-to-Image Generation},
  author={Kirstain, Yuval and Polyak, Adam and Singer, Uriel and Matiana, Shahbuland and Penna, Joe and Levy, Omer},
  booktitle={Advances in Neural Information Processing Systems},
  volume={36},
  pages={36652--36663},
  year={2023}
}

@article{wu2023hpsv2,
  title={Human Preference Score v2: A Solid Benchmark for Evaluating Human Preferences of Text-to-Image Synthesis},
  author={Wu, Xiaoshi and Hao, Yiming and Sun, Keqiang and Chen, Yixiong and Zhu, Feng and Zhao, Rui and Li, Hongsheng},
  journal={arXiv preprint arXiv:2306.09341},
  year={2023}
}

@inproceedings{luo2026editscore,
  title={Editscore: Unlocking online rl for image editing via high-fidelity reward modeling},
  author={Luo, Xin and Wang, Jiahao and Wu, Chenyuan and Xiao, Shitao and Jiang, Xiyan and Lian, Defu and Zhang, Jiajun and Liu, Dong and Liu, Zheng},
  booktitle={International Conference on Learning Representations},
  volume={2026},
  pages={33027--33056},
  year={2026}
}

@article{liu2026llbench,
  title={{LL-Bench}: Rethinking Low-Level Vision Evaluation in the Era of Large-Scale Generative Models},
  author={Liu, Lu and Duan, Huiyu and Zhu, Chenxin and Lu, Jintong and Jiang, Haoyun and Yang, Liu and Hu, Qiang and Zhai, Guangtao and Zhang, Xiaoyun},
  journal={arXiv preprint arXiv:2606.02535},
  year={2026}
}

@inproceedings{yang2022para,
  title={Personalized image aesthetics assessment with rich attributes},
  author={Yang, Yuzhe and Xu, Liwu and Li, Leida and Qie, Nan and Li, Yaqian and Zhang, Peng and Guo, Yandong},
  booktitle={Proceedings of the IEEE/CVF Conference on Computer Vision and Pattern Recognition},
  pages={19861--19869},
  year={2022}
}

@inproceedings{sheng2023aesclip,
  title={{AesCLIP}: Multi-Attribute Contrastive Learning for Image Aesthetics Assessment},
  author={Sheng, Xiangfei and Li, Leida and Chen, Pengfei and Wu, Jinjian and Dong, Weisheng and Yang, Yuzhe and Xu, Liwu and Li, Yaqian and Shi, Guangming},
  booktitle={Proceedings of the 31st ACM International Conference on Multimedia},
  pages={1117--1126},
  year={2023},
}

@inproceedings{behrad2025charm,
  title={{Charm}: The Missing Piece in {ViT} Fine-Tuning for Image Aesthetic Assessment},
  author={Behrad, Fatemeh and Tuytelaars, Tinne and Wagemans, Johan},
  booktitle={Proceedings of the IEEE/CVF Conference on Computer Vision and Pattern Recognition},
  pages={7815--7824},
  year={2025}
}

@article{liao2025humanaesexpert,
  title={{HumanAesExpert}: Advancing a Multi-Modality Foundation Model for Human Image Aesthetic Assessment},
  author={Liao, Zhichao and Liu, Xiaokun and Qin, Wenyu and Li, Qingyu and Wang, Qiulin and Wan, Pengfei and Zhang, Di and Zeng, Long and Feng, Pingfa},
  journal={arXiv preprint arXiv:2503.23907},
  year={2025}
}

@inproceedings{cao2026artimuse,
  title={{ArtiMuse}: Fine-Grained Image Aesthetics Assessment with Joint Scoring and Expert-Level Understanding},
  author={Cao, Shuo and Ma, Nan and Li, Jiayang and Li, Xiaohui and Shao, Lihao and Zhu, Kaiwen and Zhou, Yu and Pu, Yuandong and Wu, Jiarui and Wang, Jiaquan and Qu, Bo and Wang, Wenhai and Qiao, Yu and Yao, Dajuin and Liu, Yihao},
  booktitle={Proceedings of the IEEE/CVF Conference on Computer Vision and Pattern Recognition},
  pages={15313--15322},
  year={2026}
}

@article{sha2026portraitcraft,
  title={{PortraitCraft}: A Benchmark for Portrait Composition Understanding and Generation},
  author={Sha, Yuyang and Lou, Zijie and Tang, Youyun and Qu, Xiaochao and Qu, Zheng and Xia, Ben and Li, Haoxiang and Liu, Ting and Liu, Luoqi},
  journal={arXiv preprint arXiv:2604.03611},
  year={2026}
}

@article{bradley1952rank,
  title={Rank analysis of incomplete block designs: I. the method of paired comparisons},
  author={Bradley, Ralph Allan and Terry, Milton E},
  journal={Biometrika},
  volume={39},
  number={3/4},
  pages={324--345},
  year={1952}
}

@article{davidson1970ties,
  title={On extending the Bradley-Terry model to accommodate ties in paired comparison experiments},
  author={Davidson, Roger R},
  journal={Journal of the American Statistical Association},
  volume={65},
  number={329},
  pages={317--328},
  year={1970}
  }

@misc{openai2026gpt55,
  author       = {{OpenAI}},
  title        = {{GPT-5.5} System Card},
  year         = {2026},
  month        = apr,
  howpublished = {\url{https://openai.com/index/gpt-5-5-system-card/}},
}

@misc{qwen2026qwen35,
  author       = {{Qwen Team}},
  title        = {{Qwen3.5}: Towards Native Multimodal Agents},
  year         = {2026},
  month        = feb,
  howpublished = {\url{https://qwen.ai/blog?id=qwen3.5}},
}

@misc{qwen2026qwen36,
  author       = {{Qwen Team}},
  title        = {{Qwen3.6-27B}: Flagship-Level Coding in a {27B} Dense Model},
  year         = {2026},
  month        = apr,
  howpublished = {\url{https://qwen.ai/blog?id=qwen3.6-27b}},
}

@misc{qwen2026qwen37plus,
  author       = {{Qwen Team}},
  title        = {{Qwen3.7-Plus}: Multimodal Agent Intelligence},
  year         = {2026},
  month        = jun,
  howpublished = {\url{https://qwen.ai/blog?id=qwen3.7-plus}},
}

@misc{google2026gemini35,
  author       = {{Google DeepMind}},
  title        = {{Gemini 3.5 Flash} Model Card},
  year         = {2026},
  month        = may,
  howpublished = {\url{https://deepmind.google/models/model-cards/gemini-3-5-flash/}},
  note         = {Accessed: August 2, 2026}
}

@misc{anthropic2026claude48,
  author       = {{Anthropic}},
  title        = {{Claude Opus 4.8} System Card},
  year         = {2026},
  month        = may,
  howpublished = {\url{https://www.anthropic.com/system-cards}},
  note         = {Released May 28, 2026; accessed August 2, 2026}
}

@article{seed2026seed20,
  author  = {{ByteDance Seed}},
  title   = {{Seed2.0} Model Card: Towards Intelligence Frontier for
             Real-World Complexity},
  journal = {arXiv preprint arXiv:2607.00248},
  year    = {2026}
}

@article{team2026kimi,
  author  = {{Kimi Team} and Bai, Tongtong and Bai, Yifan and Bao, Yiping
             and Cai, S. H. and Cao, Yuan and Charles, Y. and Che, H. S.
             and Chen, Cheng and Chen, Guanduo and others},
  title   = {{Kimi K2.5}: Visual Agentic Intelligence},
  journal = {arXiv preprint arXiv:2602.02276},
  year    = {2026}
}

@misc{xiaomi2026mimov25,
  author       = {{Xiaomi MiMo Team}},
  title        = {{MiMo-V2.5}},
  year         = {2026},
  month        = apr,
  howpublished = {\url{https://huggingface.co/XiaomiMiMo/MiMo-V2.5}},
  note         = {Model card, accessed August 2, 2026}
}

@inproceedings{murray2012ava,
  title={{AVA}: A Large-Scale Database for Aesthetic Visual Analysis},
  author={Murray, Naila and Marchesotti, Luca and Perronnin, Florent},
  booktitle={Proceedings of the IEEE Conference on Computer Vision and Pattern Recognition},
  pages={2408--2415},
  year={2012}
}

@article{lee2018kupcp,
  title={Photographic Composition Classification and Dominant Geometric Element Detection for Outdoor Scenes},
  author={Lee, Jun-Tae and Kim, Han-Ul and Lee, Chul and Kim, Chang-Su},
  journal={Journal of Visual Communication and Image Representation},
  volume={55},
  pages={91--105},
  year={2018},
  doi={10.1016/j.jvcir.2018.05.018}
}

@inproceedings{he2023eat,
  title={{EAT}: An Enhancer for Aesthetics-Oriented Transformers},
  author={He, Shuai and Ming, Anlong and Zheng, Shuntian and Zhong, Haobin and Ma, Huadong},
  booktitle={Proceedings of the 31st ACM International Conference on Multimedia},
  pages={1023--1032},
  year={2023},
  doi={10.1145/3581783.3611881}
}

@inproceedings{cao2026unipercept,
  title={{UniPercept}: Towards Unified Perceptual-Level Image Understanding across Aesthetics, Quality, Structure, and Texture},
  author={Cao, Shuo and Li, Jiayang and Li, Xiaohui and Pu, Yuandong and Zhu, Kaiwen and Gao, Yuanting and Luo, Siqi and Xin, Yi and Qin, Qi and Zhou, Yu and Chen, Xiangyu and Zhang, Wenlong and Fu, Bin and Qiao, Yu and Liu, Yihao},
  booktitle={Proceedings of the International Conference on Machine Learning},
  year={2026}
}

@article{shao2024deepseekmath,
  title={Deepseekmath: Pushing the limits of mathematical reasoning in open language models},
  author={Shao, Zhihong and Wang, Peiyi and Zhu, Qihao and Xu, Runxin and Song, Junxiao and Bi, Xiao and Zhang, Haowei and Zhang, Mingchuan and Li, YK and Wu, Yang and others},
  journal={arXiv preprint arXiv:2402.03300},
  year={2024}
}

@article{kong2026shotcrop,
  title = {{ShotCrop}$^{3}$: Cropping Human-Centric Images into Cinematic Triple-Shot Compositions},
  author = {Kong, Dehong and Lei, Lina and Zheng, Lingtao and Wu, Chenyang and Zhang, Ailing and Qin, Xinran and Ma, Teng and Xu, Jiaqi and Wang, Zhixin and Chen, Zhikai and others},
  journal = {arXiv preprint arXiv:2606.05635},
  year = {2026}
}

\end{document}